\documentclass[lettersize,journal]{IEEEtran}
\usepackage{amsmath,amsfonts}
\usepackage{algorithmic}
\usepackage{algorithm}
\usepackage{array}
\usepackage{caption}
\usepackage{tabularray}
\usepackage{tabularx}
\usepackage{subfig}
\usepackage[export]{adjustbox}
\usepackage{svg}
\usepackage{amsmath}
\usepackage{mathrsfs}
\usepackage{textcomp}
\usepackage{stfloats}
\usepackage{tabularx,multicol,multirow}
\usepackage{url}
\usepackage{verbatim}
\usepackage{booktabs}
\usepackage{graphicx}
\usepackage{cite}
\usepackage{bm}
\usepackage{hyperref}
\hypersetup{
    colorlinks=true,
    linkcolor=black,  
    citecolor=black,  
    urlcolor=black    
}
\definecolor{purple}{rgb}{0.65,0,0.65}
\definecolor{black}{rgb}{0.0, 0.0, 0.0}
\newcommand{\chen}[1]
{{\color{black}#1}}

\begin{document}

\bibliographystyle{unsrt}
\title{Natias: Neuron Attribution based Transferable Image Adversarial Steganography}

\author{Zexin Fan, Kejiang Chen, Kai Zeng, Jiansong Zhang, Weiming Zhang, Nenghai Yu
\thanks{This work was supported in part by the Natural Science Foundation of China under Grant 62102386, 62002334, 62072421 and 62121002.}%
\thanks{All the authors are with CAS Key Laboratory of Electromagnetic Space Information, Anhui Province Key Laboratory of Digital Security, School of Information Science and Technology, University of Science and Technology of China, Hefei 230026, China.}
\thanks{Corresponding author: Kejiang Chen (Email:chenkj@ustc.edu.cn).}
}

\markboth{}%
{Shell \MakeLowercase{\textit{\textit}}: A Sample Article Using IEEEtran.cls for IEEE Journals}

\maketitle

\begin{abstract}
Image steganography is a technique to conceal secret messages within digital images. Steganalysis, on the contrary, aims to detect the presence of secret messages within images.
Recently, deep-learning-based steganalysis methods have achieved excellent detection performance.
As a countermeasure, adversarial steganography has garnered considerable attention due to its ability to effectively deceive deep-learning-based steganalysis. However, steganalysts often employ unknown steganalytic models for detection. Therefore, the ability of adversarial steganography to deceive non-target steganalytic models, known as transferability, becomes especially important. Nevertheless, existing adversarial steganographic methods do not consider how to enhance transferability. To address this issue, we propose a novel adversarial steganographic scheme named Natias. Specifically, we first attribute the output of a steganalytic model to each neuron in the target middle layer to identify critical features. Next, we corrupt these critical features that may be adopted by diverse steganalytic models. Consequently, it can promote the transferability of adversarial steganography. Our proposed method can be seamlessly integrated with existing adversarial steganography frameworks.
Thorough experimental analyses affirm that our proposed technique possesses improved transferability when contrasted with former approaches, and it attains heightened security in retraining scenarios.

\end{abstract}


\section{Introduction}\label{intro}
\IEEEPARstart{I}{mage} steganography~\cite{fridrich2009steganography,li2011survey,qzx}, as a technique of concealing secret messages within images without arousing the attention of adversaries, has garnered widespread attention in the academic community in recent years due to its significance in cybersecurity. The cover image is commonly used to denote the image without hidden secret messages, while the stego image denotes the image containing concealed messages. The most effective steganographic framework is currently the distortion minimization (DM) framework~\cite{STC}, which formalizes the steganography problem as a source coding problem with fidelity constraints. First, a steganographic distortion function is designed, measuring the risk of modifying each pixel or frequency coefficient; then, under the premise of minimizing distortion, the steganographic code is utilized for message embedding. Since the Syndrome Trellis Codes (STCs)~\cite{STC}, and Steganographic Polar Codes (SPCs)~\cite{SPC} codes have achieved performances close to the rate-distortion bound, the current focus of steganography research is on designing steganographic distortion functions. In recent years, various heuristically designed distortion functions have been proposed, such as SUNIWARD~\cite{UNIWARD}, HILL~\cite{HILL} and MiPOD~\cite{MiPOD}. In recent years, with the development of deep learning, several methods utilizing deep learning techniques for distortion learning have also been proposed, such as UT-GAN~\cite{UT-GAN}, SPAR-RL~\cite{SPAR-RL} and JoCoP~\cite{JoCoP}.

\begin{figure}[!t]
\centering
\includegraphics[width=0.99\linewidth]{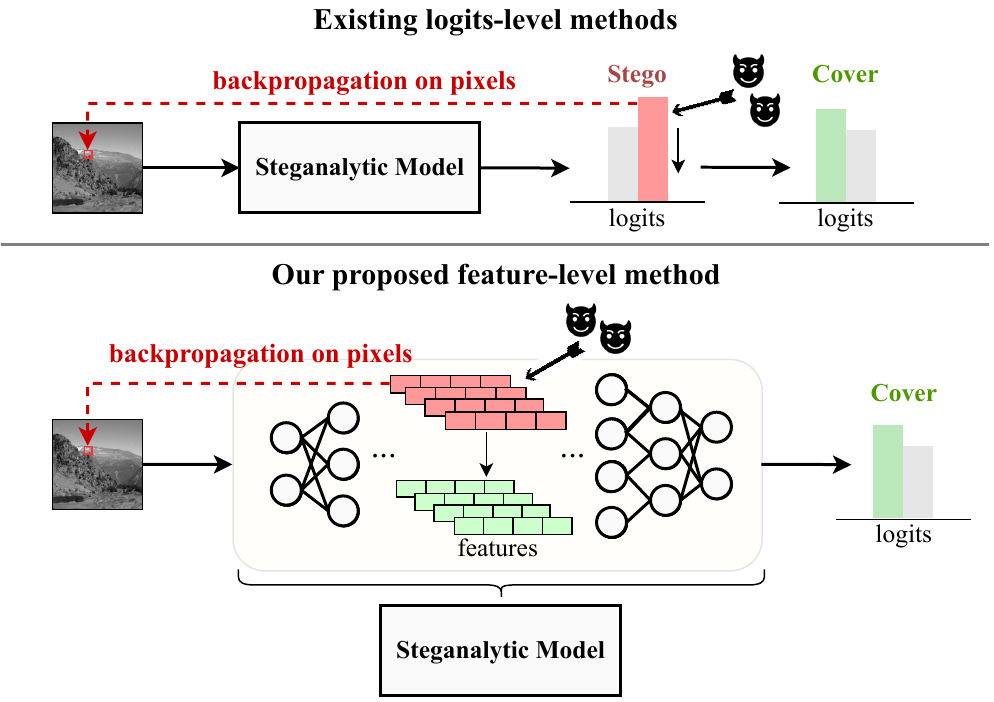}
\caption{Comparison of workflows for existing logits-level adversarial steganographic methods and our proposed feature-level adversarial steganographic method.}
\label{comparison}
\end{figure}
As the opposing side in this game, steganalysis~\cite{xia2014steganalysis} aims to detect the presence of secret messages within images. Early steganalytic methods are based on handcrafted features and divided into two stages: first, extract handcrafted features such as SRM~\cite{SRM}, DCTR~\cite{DCTR}, and GFR~\cite{GFR} from the image to be detected, and then use machine learning tools such as Support Vector Machines (SVMs) and ensemble classifiers to classify the extracted features. The two-step operation of these methods is challenging to optimize simultaneously, thus limiting their performance. In recent years, with the rapid advancement of deep learning, a variety of steganalytic methods~\cite{boroumand2018deep,you2020siamese} based on convolutional neural networks (CNNs) have emerged. These modern steganalytic models, including SRNet~\cite{SRNet}, SiaStegNet~\cite{SiaStegNet}, CovNet~\cite{CovNet}, and LWENet~\cite{LWENet}, utilize CNN to automatically extract discriminative features, effectively enhancing the accuracy of steganalysis. Therefore, these CNN-based steganalytic methods have presented great challenges for steganographers. 

To cope with this challenge, steganographers leverage the vulnerability of deep learning classification models, i.e., adversarial examples~\cite{szegedy2013intriguing,goodfellow2014explaining}, to deceive CNN-based steganalytic models.
Specifically, an adversarial example involves introducing a subtle and imperceptible perturbation to the image to deceive the classification models.
From this perspective, a set of methods has emerged, which can be categorized into three groups: cover enhancement based methods, distortion adjustment based methods and stego post-processing based methods. 

Cover enhancement methods aim to generate adversarial cover images that can be identified as the initial covers, even when steganographic modifications are applied. Common cover enhancement based methods include ADS~\cite{ADS} and SPS-ENH~\cite{SPS-ENH}.  Distortion adjustment based methods, such as ADV-EMB~\cite{ADV-EMB}, AEN~\cite{AEN}, CR-AIS~\cite{CRAIS}, Backpack~\cite{backpack}, Min-max~\cite{min_max} and JS-IAE~\cite{JS-IAE}, improve the priority of modification directions that can deceive steganalytic models by adjusting distortion within the DM framework. Stego post-processing based methods, including ECN~\cite{zha_ECN}, ISteg~\cite{ISteg}, and USGS~\cite{stego_selection}, add adversarial perturbations to stego images or filter stego images using statistical features to enhance security without affecting the extraction of messages. These adversarial steganographic methods all aim to deceive several certain
target steganalytic models from the logits level. Consequently, it is easy to overfit the target model, leading to a lack of transferability and ultimately causing failure to deceive other non-target steganalytic models.

According to Kerckhoffs's principle, however, in practice, steganalysts do not use the \chen{pre-defined} target model steganographers attacking for detection, instead, they use different steganalytic models. In other words, there is asymmetry between the steganographer and the steganalyst. Regardless of the model chosen by the steganographer as the target steganalytic model, the steganalyst always has the flexibility to choose different models to render the adversarial attack ineffective. For example, the steganalyst may use a model with different parameters or structures from the one attacked by the steganographer, or even employ entirely different detection methods. Therefore, it is essential for adversarial steganography to exhibit adequate transferability, i.e., the ability to effectively deceive non-target steganalytic models.

\begin{figure*}[!t]
\centering
\subfloat[A cover example]{\includegraphics[width=0.23\linewidth,height=0.23\linewidth, frame]{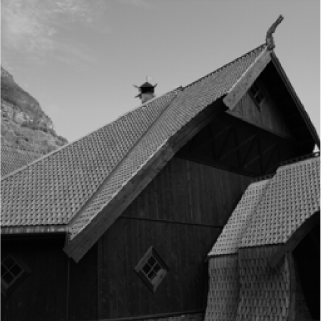}}
\hfil
\subfloat[SRNet's attention distribution]{\includegraphics[width=0.23\linewidth,height=0.23\linewidth]{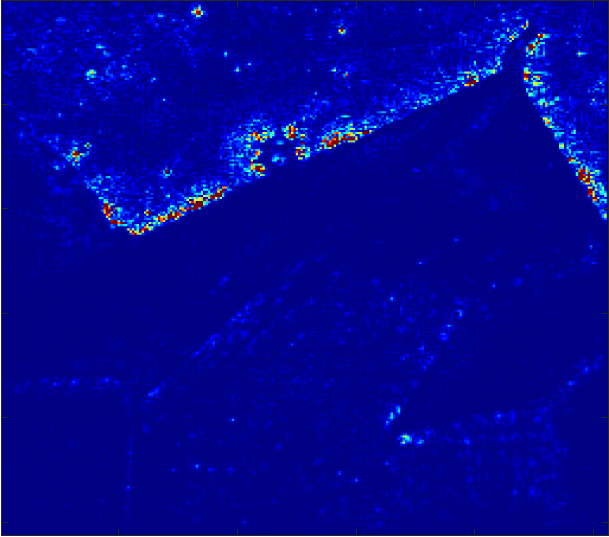}}
\hfil
\subfloat[CovNet's attention distribution]{\includegraphics[width=0.23\linewidth,height=0.23\linewidth]{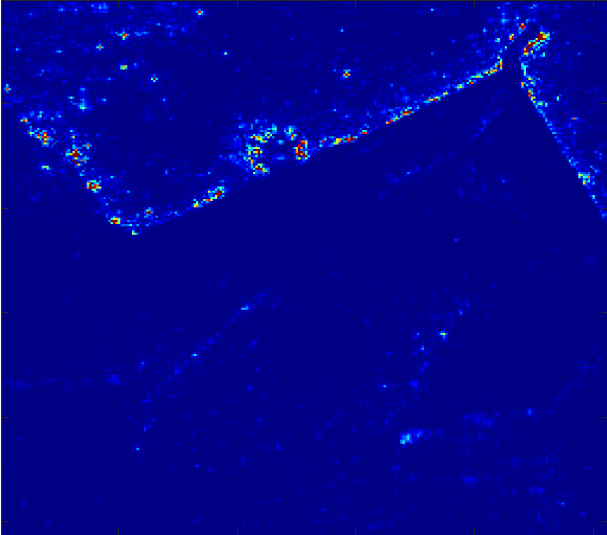}}
\hfil
\subfloat[LWENet's attention distribution]{\includegraphics[width=0.23\linewidth,height=0.23\linewidth]{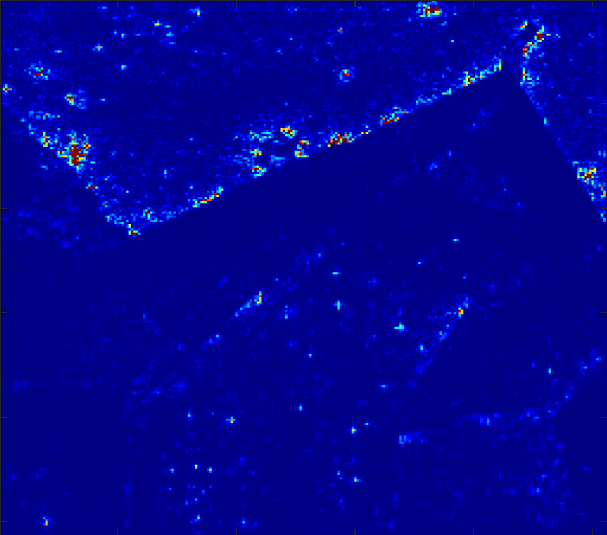}}
\\
\subfloat[Corresponding stego]{\includegraphics[width=0.23\linewidth,height=0.23\linewidth, frame]{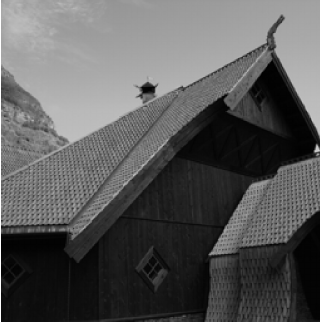}}
\hfil
\subfloat[SRNet's attention distribution]{\includegraphics[width=0.23\linewidth,height=0.23\linewidth]{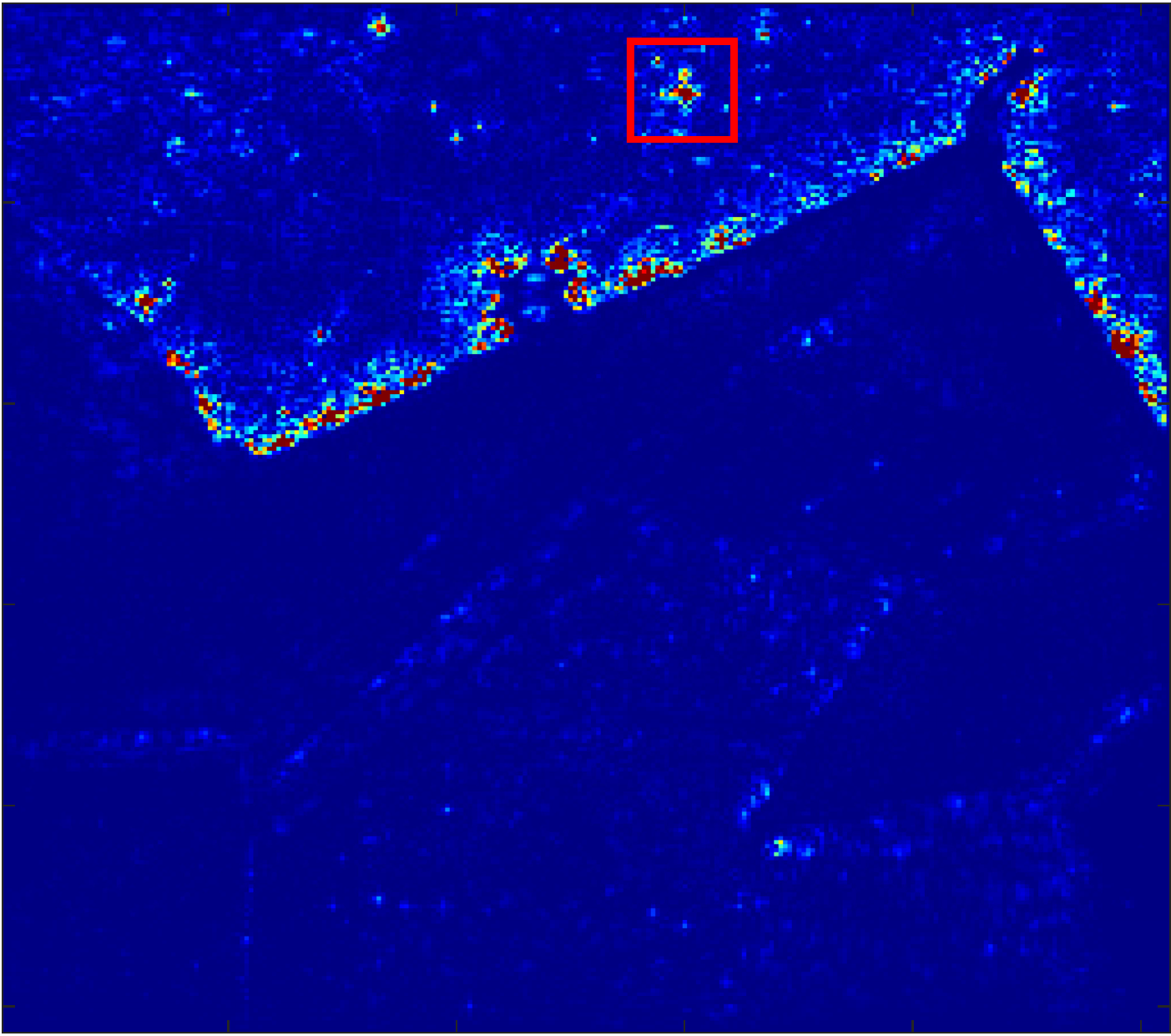}}
\hfil
\subfloat[CovNet's attention distribution]{\includegraphics[width=0.23\linewidth,height=0.23\linewidth]{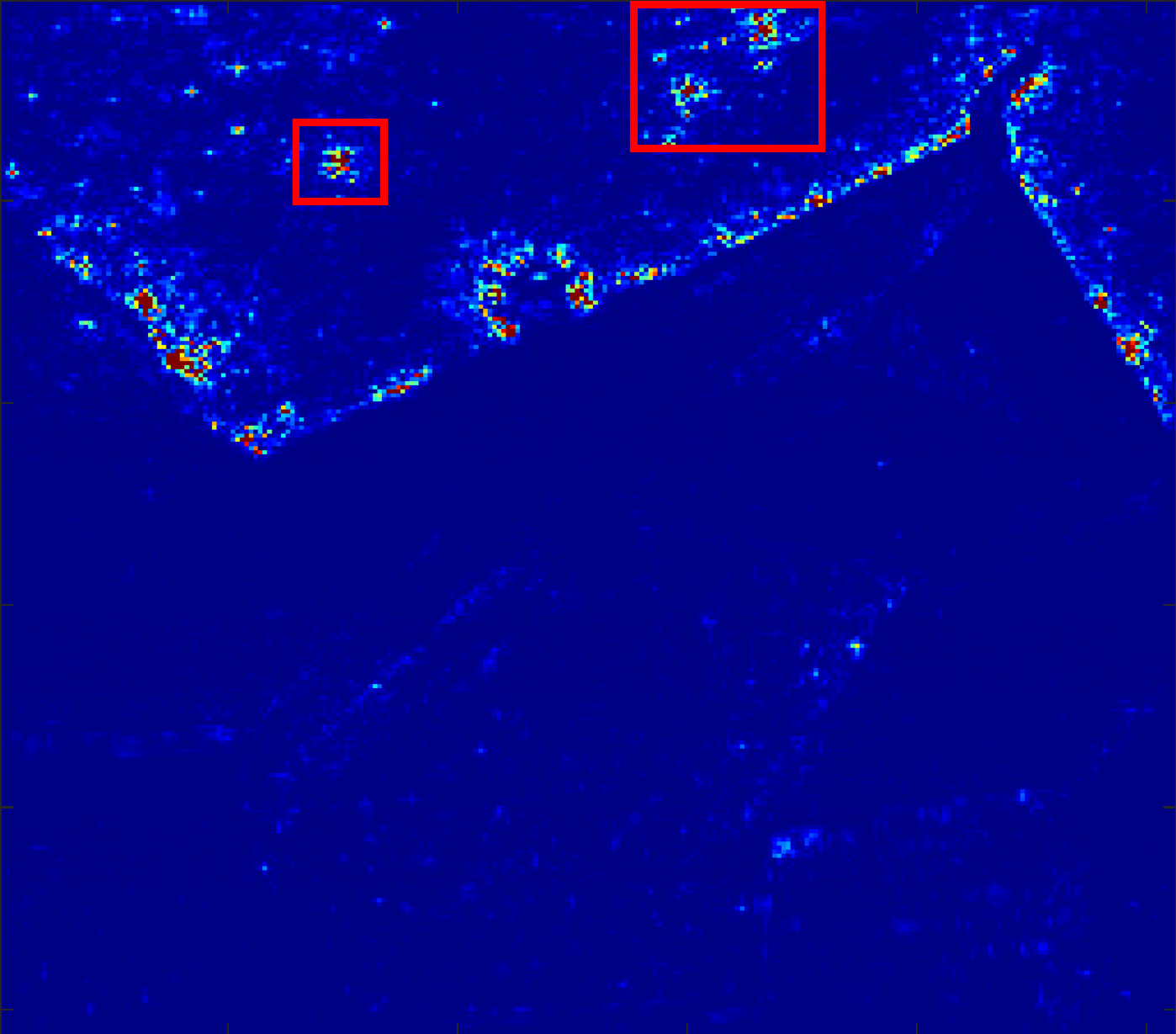}}
\hfil
\subfloat[LWENet's attention distribution]{\includegraphics[width=0.23\linewidth,height=0.23\linewidth]{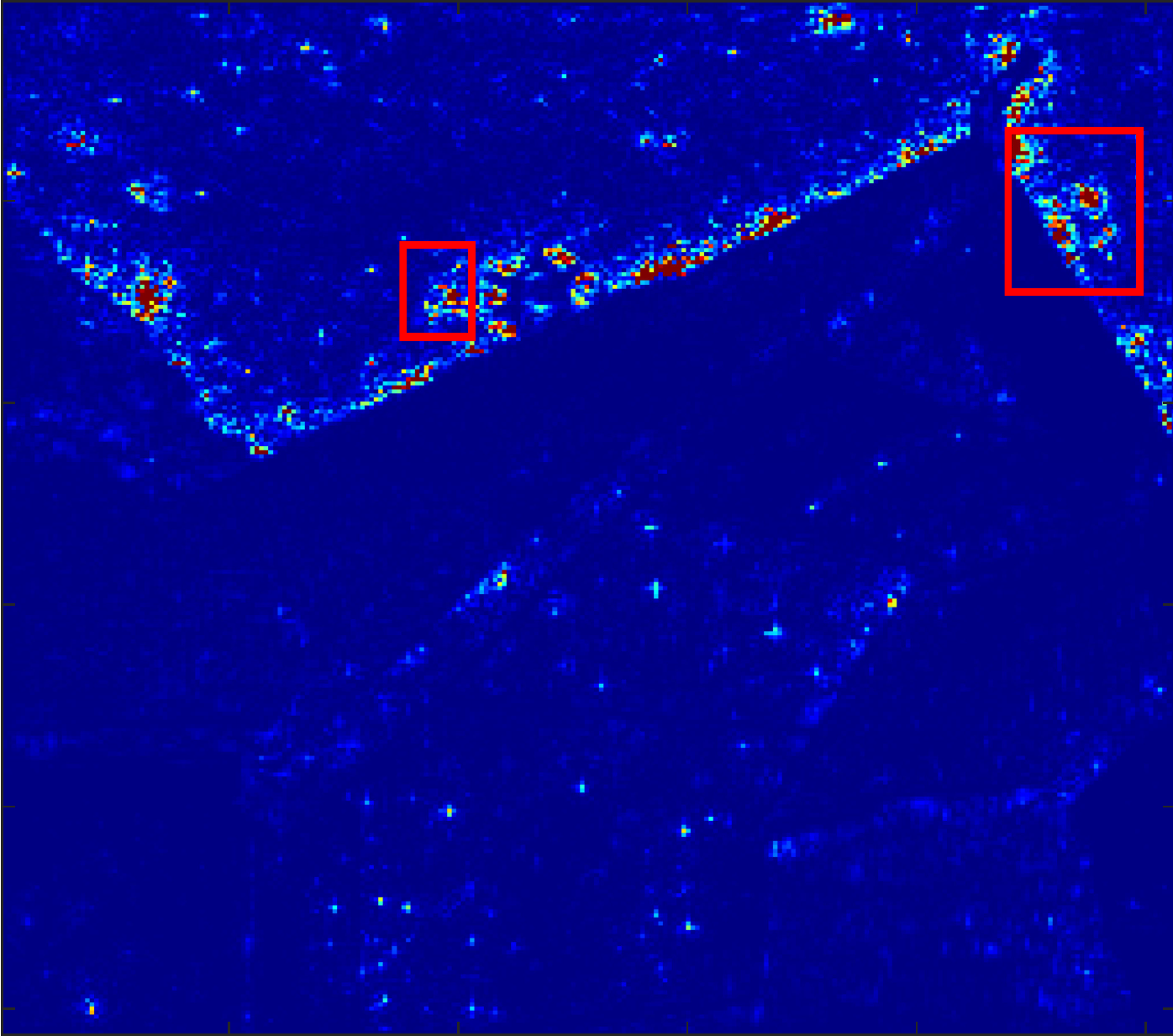}}
\caption{Visualization of the attention distributions of three steganalytic models (SRNet, CovNet, and LWENet). The redder regions possess higher importance to the decision of the steganalytic model. The top row shows attention distributions of different steganalytic models when detecting the cover. The bottom row shows attention distributions of different steganalytic models when detecting the corresponding stego, which is generated by using our proposed Natias to attack CovNet. The regions enclosed by the red rectangles denote notable alterations in the attention maps.}
\label{attention_map}
\end{figure*}

To address this issue, we propose a novel adversarial steganographic scheme named Natias. As depicted in Fig.~\ref{comparison}, our approach differs from previous logits-level adversarial steganographic methods as it launches attacks at the feature level. Inspired by~\cite{CNN_feature}, we infer that different classifiers performing the same classification task often rely on significantly overlapping features, referred to as critical features.
Our observations also support this inference, where different steganalytic models indeed allocate greater attention to image patterns according to these critical features when detecting the same image. For instance, as illustrated in Fig.~\ref{attention_map}, when diverse steganalytic models analyze the same cover image, their attention distributions exhibit notable similarity (in this example, their points of focus are concentrated on the edge regions of the object). Following this line of thought, we first employ neural network attribution techniques to characterize the importance of different intermediate layer features. Subsequently, we expose the critical features of the intermediate layer to adversarial attacks. Finally, we use the gradient map obtained to enhance steganographic security combined with existing adversarial steganographic schemes. Our main contributions are summarized as follows:
\begin{itemize}
    \item We propose \textbf{Natias} to enhance adversarial steganography transferability by attacking the intermediate layer features of the steganalytic model. The integrated gradients attribution method, which effectively adapts to the subtle nature of steganographic signals and mitigates the issue of gradient vanishing, is utilized to identify critical features.
    \item Our proposed method can be seamlessly integrated with existing adversarial steganography frameworks, including cover enhancement based methods such as SPS-ENH, distortion adjustment based methods such as ADV-EMB, and stego post-processing based methods such as USGS.
    \item Extensive experimental results show that our proposed method can achieve state-of-the-art transferability while maintaining high performance against the target steganalytic model and comparable security performance in the retraining scenario.
\end{itemize}

The rest of this paper is arranged as follows. Sec.~\ref{related} gives a brief overview of related steganographic methods based on adversarial examples and some other works. Sec.~\ref{method} encompasses a comprehensive explanation of the proposed method. Sec.~\ref{experiments} conducts analysis and discussions of experimental results. Finally, the overall conclusion of this paper along with prospects for future work are given in Sec.~\ref{conclusion}.

The source code of our implementations of Natias can be found at \href{https://github.com/Van-ZX/Natias}{\textcolor[RGB]{83, 168, 236}{https://github.com/Van-ZX/Natias}}.

\section{Preliminaries}\label{related}

In this section, we introduce some concepts and review related work on adversarial steganography that will be used in the following sections.

\subsection{Steganalytic Model}
The steganalytic model is essentially a binary classifier designed to differentiate stego images from cover images. Let $\mathscr{F}$ represent the steganalytic model, where its input is an image $\bm{x}$ (cover image or stego image), and obtain the decision criterion as follows:
\begin{equation}
    \mathscr{F}(\bm{x})=\left\{\begin{array}{lc}
        0, & \text{ if }\Phi(\bm{x})<0.5\\
        1, & \text{ if }\Phi(\bm{x})\ge0.5,
        \end{array}\right.
\end{equation}
where $\Phi(\bm{x})\in[0, 1]$ indicates that the probability steganalytic model regards the input image $\bm{x}$ as a stego image. $\mathscr{F}=0$ implies that $\bm{x}$ is a cover image, while $\mathscr{F}=1$ implies that $\bm{x}$ is a stego image. To assess the security of steganographic algorithms, we introduce two metrics: the missed detection rate $P_\text{MD}$ and the false alarm rate $P_\text{FA}$. The missed detection occurs when stego images are misclassified, and the false alarm occurs when cover images are misclassified. The corresponding error probabilities are defined as follows:
\begin{equation}
    P_\text{MD}=\text{Pr}\{\mathscr{F}(\bm{x})=1|\bm{x}\in \mathcal{S}\},
\end{equation}
\begin{equation}
    P_\text{FA}=\text{Pr}\{\mathscr{F}(\bm{x})=0|\bm{x}\in \mathcal{C}\},
\end{equation}
where $\mathcal{C}$ and $\mathcal{S}$ are the cover set and stego set respectively. With an equal Bayesian prior for cover and stego images, the total error rate is:
\begin{equation}
P_\text{E}=\frac{P_\text{MD}+P_\text{FA}}{2}.
\end{equation}

Alternatively, the performance of steganalytic models can be evaluated using detection accuracy:
\begin{equation}
    Acc = 1-P_\text{E}.
\end{equation}

\subsection{Adversarial Steganography and Typical Methods}
In recent years, several steganographic methods based on adversarial examples have been proposed. These methods initially train a steganalytic model $\mathscr{F}$ based on existing steganographic algorithms to be enhanced. 
Subsequently, an image $\bm{x}$ is fed into the pre-trained steganalytic model as input, and then, we can obtain a gradient map $\bm{G}_{\bm{x}}$ according to the given loss function 
$L(\bm{x},t;\mathscr{F})$ to enhance the security of the steganographic algorithm, which is defined as follows:
\begin{equation}
    \bm{G}_{\bm{x}}  =\nabla_{\bm{x}} L\left(\bm{x}, t ; \mathscr{F}\right),
\end{equation}
where $t$ is the target label ($t=0$ denotes cover, $t=1$ denotes stego), $L(\bm{x},t;\mathscr{F})$ is the loss function of $\mathscr{F}$, and $\nabla_{\bm{x}}$ represents the partial derivative with respect to the input $\bm{x}$.

Adversarial steganography can be broadly classified into the following three categories, based on the different ways of utilizing gradients:

\subsubsection{Cover Enhancement Based Methods}
Cover enhancement based methods involve modifying the cover according to gradients to create an enhanced cover that can withstand steganalysis detection even if steganographic noise is added. Among these methods, sparse enhancement (SPS-ENH)~\cite{SPS-ENH} is the most representative, and its main process can be described as follows.:
\begin{itemize}
    \item [(1)]
    Train a CNN-based steganalytic model $\mathscr{F}$ according to the cover set and corresponding stego set created by using the existing distortion function.
    \item [(2)] 
    For a given cover $\bm{c}$, the corresponding stego is input into the steganalytic model $\mathscr{F}$ to generate the gradient map $\bm{G}$. Leverage a mask $\bm{m}$ to control whether elements can be enhanced and $\bm{m}=\bm{0}$ in the initial state. Let $\bm{G}'=\bm{G}\cdot(\bm{1}-\bm{m})$.
    \item [(3)]
    Select the top $k$ elements from $\bm{G}'$ and set the corresponding positions in $\bm{m}$ to 1. Let $\bm{e}=\bm{c}+\bm{G}'$ be the enhanced cover, and the corresponding stego is $\bm{s}$. When the probabilistic output of $\bm{s}$ is larger than a threshold $\tau$, scramble the secret message and regenerate the stego.
    \item [(4)]
    Repeat steps (2) and (3) until the adversarial stego image is predicted as the cover. Otherwise, the message will be embedded according to the initial distortion function to obtain the final stego.
\end{itemize}

Cover enhancement methods directly manipulate the cover image based on gradients to generate an enhanced cover that is more effective in deceiving steganalytic models. Consequently, even when steganographic noise is added to these enhanced covers, they can still avoid being detected. However, this kind of method may introduce unforeseen artifacts in the cover, thereby disrupting the statistical characteristics of the cover image. The disruption makes it possible to detect the corresponding stego images using traditional handcrafted feature based classifiers and non-target steganalytic models.

\subsubsection{Distortion Adjustment based Methods}
Distortion adjustment based methods utilize gradients to adjust the distortion value of existing distortion functions within the DM framework. This refinement encourages steganographic modifications at locations that enhance the concealment of the resulting stego image to avoid being detected by steganalytic models. Adversarial embedding (ADV-EMB)~\cite{ADV-EMB} is one of the most classic distortion adjustment based methods and its principal workflow is outlined as follows:
\begin{itemize}
    \item [(1)]
    Train a CNN-based steganalytic model $\mathscr{F}$. For each cover $\bm{c}$, compute the initial embedding distortion $\{\rho_i^+,\rho_i^-\}$ for elements, where $\rho_i^+$ and $\rho_i^-$ represent the distortion of $+1$ and $-1$ on the $i$-th corresponding element. Initialize the step parameter $\beta$.
    \item [(2)] 
    Divide the elements in $\bm{c}$ into two disjoint groups: a common group and an adjustable group containing $1-\beta$ and $\beta$ of embedding units separately. Embed $1-\beta$ of the secret message into the common group according to the initial distortion and then obtain the corresponding gradients $\bm{G}_{\bm{c}}=\{g_i\}_{i=1}^n$ of $\mathscr{F}$ with respect to it.
    \item [(3)] 
    Adjust the distortion of the adjustable group based on the sign of $g_i$ as follows:
    \begin{equation}
        \begin{array}{l}
        \rho_{a d v+}^{i}=\left\{\begin{array}{ll}
        \rho_i^+ / \alpha, & g_i<0 \\
        \rho_i^+, & g_i=0 \\
        \rho_i^+ * \alpha, & g_i>0
        \end{array}\right. \\
        \rho_{a d v-}^{i}=\left\{\begin{array}{ll}
        \rho_i^- / \alpha, & g_i>0 \\
        \rho_i^-, & g_i=0 \\
        \rho_i^- * \alpha, & g_i<0
        \end{array}\right.
        \end{array},
    \end{equation}
    where $\alpha$ is a scaling factor larger than $1$. And then embed the rest $\beta$ secret messages according to the adjusted distortion $\{\rho_{a d v+}^{i},\rho_{a d v-}^{i}\}$.
    \item [(4)]
    Update parameter $\beta=\beta+0.1$ and repeat steps (2) and (3) until the corresponding stego of $\bm{c}$ can deceive $\mathscr{F}$.
\end{itemize}

There is another specific distortion adjustment based method called Backpack~\cite{backpack}. Unlike previous methods, Backpack does not calculate gradients for the input image. Instead, it directly calculates derivatives with respect to the distortion and then updates the distortion using gradient descent:
\begin{equation}
    \bm{\rho}\leftarrow\bm{\rho}-\alpha\nabla_{\bm{\rho}} L(\bm{x},t;\mathscr{F}),
\end{equation}
\begin{equation}
    \nabla_{\bm{\rho}} L(\bm{x},t;\mathscr{F})=\nabla_{\bm{x}} L\cdot\frac{\partial \bm{x}}{\partial \bm{\pi}}\cdot(\frac{\partial \bm{\pi}}{\partial\bm{\rho}}-\frac{\partial \bm{\pi}}{\lambda}(\frac{\partial H(\bm{\pi})}{\partial\lambda})^{-1}\nabla_{\bm{\rho}} H(\bm{\pi})),
\end{equation}
where $H(\bm{\pi})=-\sum_{i=1}^n\pi_ilog\pi_i$ represents the information entropy of the modification probability, and the meanings of the remaining parameters are consistent with the above.

Additionally, Min-Max~\cite{min_max} is also a classic steganographic scheme. It models the adversary-aware case as a sequential min-max game, where Alice is the steganographer and Eve is the steganalyst. Initially, Eve trains a steganalytic model based on the initial dataset. Subsequently, Alice employs ADV-EMB to create adversarial stego images, from which, Eve will select the least detectable stego images and further train a more proficient steganalytic model. Finally, Alice uses ADV-EMB to attack the newly improved steganalytic model to create even more challenging adversarial stego images. This iterative procedure persists until the created stego images remain undetectable by the target classifier. Therefore, Min-Max can be considered as the iterative version of ADV-EMB

In contrast to the cover enhancement based methods, this methodology achieves a lower level of over-adaptation. Although adversarial stego images have a slightly higher rate of modifications than conventional stego images, they are less detectable by other advanced handcrafted feature-based classifiers.

\subsubsection{Stego Post-processing based Methods}
In addition to the aforementioned two adversarial steganographic methods, there exists another method applied after obtaining the stego image, called stego post-processing based methods.

Stego generation and selection (USGS)~\cite{stego_selection} is a representative stego post-processing based adversarial steganographic method. It refines conventional distortion adjustment based methods, which consist of three main stages: pre-training the steganalytic network, stego generation, and stego selection. First, it pre-trains a steganalytic network based on the initial stego images dataset. Subsequently, regions with significant gradients and minimized distortions are identified, and adjustments are made to the distortions corresponding to these areas based on gradients. When employing varying selection thresholds, multiple distinct stego images can be obtained. These stego images, along with the initial stego image, constitute the candidate set. Then, those stego images capable of deceiving the target steganalytic model are selected. Finally, it obtains the set $H_{\bm{c}}$ of adaptive high-pass filters for the given cover $\bm{c}$ and employs the resulting set $H_{\bm{c}}$ to calculate the corresponding image residuals of the cover image and the candidate stego images available. The image with the minimal residual between the cover is chosen as the ultimate stego image.

The three mentioned adversarial steganographic methods: cover enhancement based methods, distortion adjustment based methods, and stego post-processing based methods are applied before secret message embedding, during embedding, and after embedding, respectively. All of these approaches based on adversarial examples enhance steganographic security techniques from different perspectives. 

Nevertheless, these conventional logits-level adversarial steganographic methods are confined to deceiving the target steganalytic model and always fall short in enhancing the transferability. Consequently, they face challenges in meeting the requirements of real-world scenarios.

\section{Method}\label{method}

\begin{figure*}[!t]
\centering
\includegraphics[width=0.95\linewidth]{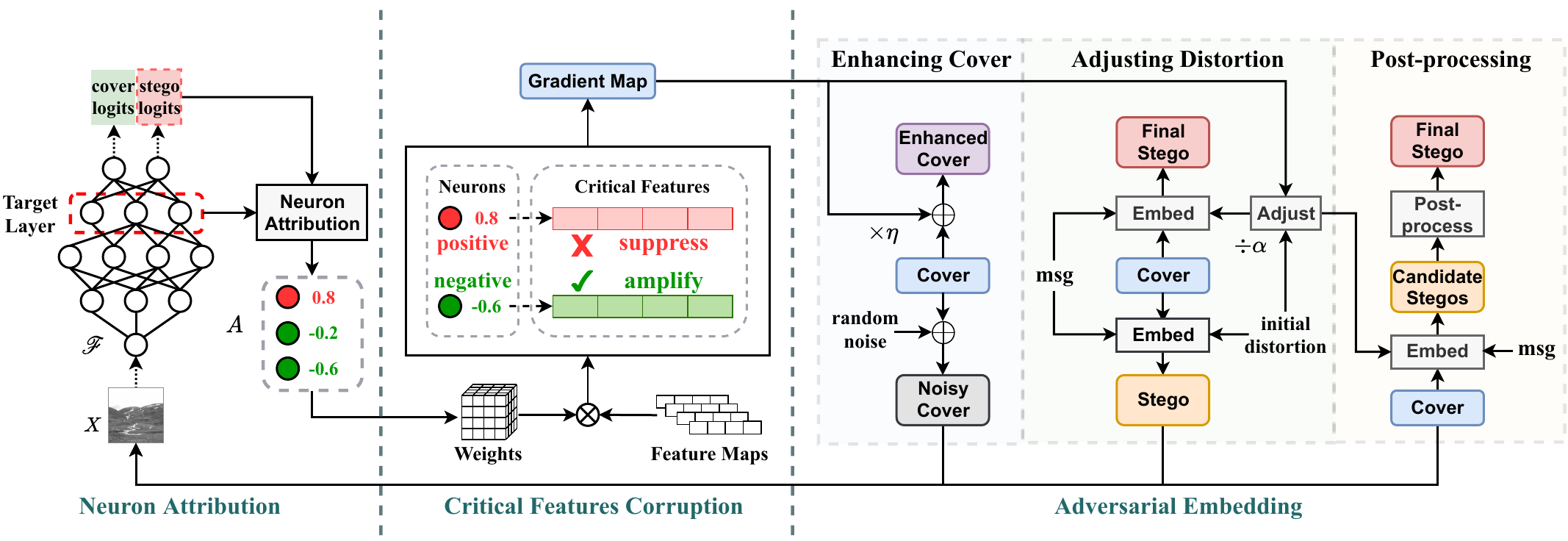}
\caption{The framework of our proposed Natias method. ``$A$" denotes the neuron attribution results, ``$\otimes$" denotes the element-wise product, ``$\oplus$" denotes the element-wise addition, ``$\eta$" denotes the coefficient controlling the magnitude of enhancing cover, ``$\alpha$" denotes the coefficient controlling the magnitude of adjusting distortion, and ``$\bm{msg}$" denotes the secret message to be embedded.}
\label{framework}
\end{figure*}

To address the above problem, we propose a neuron-attribution-based adversarial steganographic method called Natias to enhance the transferability of adversarial steganography. 

In general, a CNN network can be divided into two parts: a hierarchical feature extractor and a softmax classifier. Razavian~\cite{CNN_feature} pointed out that the features acquired by the CNN's feature extractor are generally generic and possess a robust ability to adapt across different domains and tasks. Inspired by this finding, we hypothesize that various classifiers, when performing the same classification task, are likely to depend on specific common critical features. Hence, if the adversarial noise we create not only misleads the classifier's final prediction but also heavily corrupts the crucial intermediate layer features that the classifier relies on, then our generated adversarial stego images can demonstrate enhanced transferability. In summary, the core idea of our approach is to utilize attribution techniques to extract crucial neurons, thereby delineating critical features that different steganalytic models rely on when making decisions. 

In the following, we will delve into the details of our proposed method. As illustrated in Fig.~\ref{framework}, our proposed method comprises three stages: neuron attribution, critical features corruption, and adversarial embedding. In contrast to current adversarial steganographic methods, Natias launches feature-level attacks against the target steganalytic model instead of logits-level attacks.

\subsection{Neuron Attribution}
First, we need to select a target layer to launch adversarial attacks on intermediate layer features. In our method, we choose a relatively narrow layer as the target layer because the corresponding features are more concentrative, making it easier for us to achieve successful attacks. In Sec.~\ref{experiments}, we will experimentally discuss more details about the selection of the target layer.

In addition, the crucial aspect of attacking intermediate layer features lies in identifying a suitable technique to evaluate the prominence of each neuron in the representation of features. The previous adversarial steganographic methods directly use gradients as guidance to deceive steganalytic models. However, due to the weak and sparse nature of steganographic signals, it is always difficult to capture the true impact of steganographic modifications, thus making it ineffective to deceive the target steganalytic model.

To comprehensively capture the impact of steganographic modifications on the steganalytic model and accurately evaluate the prominence of distinct neurons, we utilize integrated gradients for neuron attribution. Assume that $\bm{x}=(x_1, x_2, ..., x_n)$ represents an image with $n$ pixels, and $\bm{x}'=(x'_1, x'_2, ..., x'_n)$ denotes a baseline image with dimensions equivalent to those of $\bm{x}$. The integrated gradients of $\bm{x}$ with respect to the baseline image $\bm{x}'$ are as follows:
\begin{equation} \label{IG}
    \begin{aligned}IG:=\sum_{i=1}^n(x_i-x_i')\int_0^1\frac{\partial \mathscr{F}}{\partial x_i}(\bm{x}'+\gamma(\bm{x}-\bm{x}'))d\gamma\end{aligned},
\end{equation}
where $\mathscr{F}$ represents the steganalytic model, $\frac{\partial \mathscr{F}}{\partial x_i}$ signifies the partial derivative of $\mathscr{F}$ with respect to the $i$-th pixel, and $\gamma$ is the scaling factor. Eq.~(\ref{IG}) conveys the effect of transitioning from the baseline image $\bm{x}'$ to the image $\bm{x}$ along the straight line $\bm{x}'+\gamma(\bm{x}-\bm{x}')$ on the output of the steganalytic model $\mathscr{F}$.

However, conventionally setting the baseline image as a black image ($\bm{x}'=\bm{0}$) to compute integrated gradients does not adequately adapt to the characteristics of steganography. Because the magnitude of steganographic modifications relative to the distance from the baseline image to the input image is too minor. Therefore, considering the subtle nature of steganographic signals, to comprehensively evaluate the impact of all potential steganographic modifications, we set $\bm{x}'=\bm{x}-\bm{1}$ as the baseline image, and compute the integrated gradients along the straight line $\bm{x}+(\gamma-1)\bm{1}$, where $\gamma\in(0, 2)$. This configuration allows us to encompass all possible steganographic modifications, making the attribution result better tailored to the requirements of steganography, which can be shown in the following equation:
\begin{equation} \label{att}
    \begin{aligned}A:=\sum_{i=1}^n\int_0^2\frac{\partial \mathscr{F}}{\partial x_i}(\bm{x}+(\gamma-1)\bm{1})d\gamma\end{aligned},
\end{equation}
Eq.~(\ref{att}) describes the attribution of the output results to pixels. However, to characterize the importance of intermediate layer features, it is necessary to attribute the output results to those features. Let $\bm{y}_j$ represent the outputs of the $j$-th neuron of the target layer $\bm{y}$ in the steganalytic model, which can be regarded as intermediate layer features. Thus, similarly, we can obtain attributions of $\mathscr{F}$ to $\bm{y}_j$:

\begin{equation}\label{fea_att}
    \begin{aligned}A_{\bm{y}_j}&=\sum_{i=1}^n\int_0^2\frac{\partial \mathscr{F}}{\partial \bm{y}_j}(\bm{y}(\bm{x}_\gamma))\cdot\frac{\partial \bm{y}_j}{\partial x_i}(\bm{x}_\gamma)d\gamma
    \end{aligned}
\end{equation}
where $\bm{x}_\gamma=\bm{x}+(\gamma-1)\bm{1}$ indicates the straight line chosen for calculating the path integral and $\cdot$ represents the inner product of vectors. The attribution of $\mathscr{F}$ to the whole target layer $\bm{y}$ is $A_{\bm{y}}=\{A_{\bm{y}_j}\}_{\bm{y}_j\in\bm{y}}$. Note that $\sum_{\bm{y}_j\in y}A_{\bm{y}_j}=A$ always holds regardless of which layer we choose. Therefore, Eq.~(\ref{fea_att}) represents the importance of the feature corresponding to neuron $\bm{y}_j$. In practical implementation, we sample $M$ images along the straight line 
$\bm{x}_\gamma$ and calculate the Riemann sum to approximate the integral:
\begin{equation}\label{riemann}
    \begin{aligned}A_{\bm{y}_j}&\approx \sum_{i=1}^n\left [\frac{2}{M}\sum_{m=1}^M\frac{\partial \mathscr{F}}{\partial \bm{y}_j}(\bm{y}(\bm{x}_m))\cdot\frac{\partial \bm{y}_j}{\partial x_i}(\bm{x}_m)\right ] \\
    & = \frac{2}{M} \sum_{m=1}^{M} \left [\frac{\partial \mathscr{F}}{\partial \bm{y}_j}\left(\bm{y}\left(\bm{x}_{m}\right)\right)\cdot\sum_{i=1}^n\frac{\partial \bm{y}_j}{\partial x_i}(\bm{x}_m)\right ] \\
    & = \frac{2}{M} \sum_{m=1}^{M}\frac{\partial \mathscr{F}}{\partial \bm{y}_j}\left(\bm{y}\left(\bm{x}_{m}\right)\right) \cdot\frac{1}{M} \sum_{m=1}^{M}\sum_{i=1}^n\frac{\partial \bm{y}_j}{\partial x_i}(\bm{x}_m) \\
    & \approx (\bm{y}_j-\bm{y}_j')\cdot \frac{1}{M} \sum_{m=1}^{M}\frac{\partial \mathscr{F}}{\partial \bm{y}_j}\left(\bm{y}\left(\bm{x}_{m}\right)\right).
    \end{aligned}
\end{equation}
where $\bm{y}_j'$ represents the output of the corresponding neuron when the baseline image $\bm{x}'$ is the input. The reason why the third equality in Eq.~(\ref{riemann}) holds is that the covariance between sequences $ \{\frac{\partial \mathscr{F}}{\partial \bm{y}_j}\left(\bm{y}\left(\bm{x}_{m}\right)\right) \}_{m=1}^M$ and $ \{\sum_{i=1}^n\frac{\partial \bm{y}_j}{\partial x_i}(\bm{x}_m)\}_{m=1}^M$ is zero. Because the former represents the derivative of $\mathscr{F}$ with respect to $\bm{y}_j$, which is related to the layers after $\bm{y}$ of the steganalytic model, while the latter represents the derivative of $\bm{y}_j$ with respect to $x_i$, which is related to the layers before $\bm{y}$ of the steganalytic model.

\subsection{Critical Features Corruption}
Following the neuron attribution stage, we can obtain the magnitude of the contribution of each neuron in the target layer when the steganalytic model makes final decisions, which also represents the significance of the corresponding features associated with the neurons. In this stage, we expect the generated stego image to not only mislead the final decision of the target steganalytic model but also corrupt the critical intermediate layer features.

We categorize the neurons of the target layer into two groups based on the sign of the attribution values: positive attribution neurons and negative attribution neurons. The features corresponding to them are denoted as positive attribution features and negative attribution features, respectively. Positive attribution features strongly influence the steganalytic model toward predicting stego outcomes, whereas negative attribution features lead the steganalytic model to predict cover outcomes. 

Continuing, we suppress the impact of positive attribution features and amplify the impact of negative attribution features, thereby achieving the corruption of critical features. This manipulation guides the target steganalytic model to produce erroneous predictions. To achieve this objective, we select the neurons with attribution results whose absolute values exceed $T$ and identify the corresponding features as critical features. Then, we impose constraints on these neurons, formulating the attribution loss:
\begin{equation}\label{attloss}
    L_{att}(\bm{x}_{adv}) = \sum_{|A_{\bm{y}_j}|>T}( A_{\bm{y}_j}(\bm{x}_{adv})),
\end{equation}

On the other hand, to corrupt the critical features associated with these neurons, we utilize the aforementioned neuron attribution results as weights and multiply them with the corresponding features to design the feature loss $L_{fea}$, which can be mathematically illustrated as follows:
\begin{equation}\label{fealoss}
    L_{fea}(\bm{x}_{adv}) = \sum( A_{\bm{y}}\otimes\bm{y}(\bm{x}_{adv})),
\end{equation}
which takes the neuron attribution polarity and value magnitude into consideration, where $\bm{y}(\bm{x}_{adv})$ represents the output of the $j$-th neuron of the target layer when the input is an adversarial stego $\bm{x}_{adv}$. Therefore, the total loss is:
\begin{equation}\label{loss}
    L(\bm{x}_{adv}) = L_{att}(\bm{x}_{adv}) + \lambda L_{fea}(\bm{x}_{adv}),
\end{equation}
where $\lambda$ is the weight parameter. Finally, we can formulate the optimization problem to be solved as follows:
\begin{equation}\label{optimize}
    \text{arg }\mathop{\text{min}}\limits_{\bm{x}_{adv}}L(\bm{x}_{adv}), \text{ s.t. } \|\bm{x}_{adv}-\bm{x}\|_\infty=1
\end{equation}
Consequently, useful positive attribution features are suppressed and harmful negative attribution features are amplified.

\subsection{Adversarial Embedding}

We use the gradient descent method to solve the aforementioned problem. After obtaining the gradient map, we can utilize it for adversarial embedding. It should be noted that our proposed method can be integrated with various existing adversarial steganographic schemes.

If combined with cover enhancement based methods, such as SPS-ENH, given a cover $\bm{c}$ and a pre-trained steganalytic model $\mathscr{F}$, we first use a basic distortion function such as SUNIWARD to obtain the conventional stego $\bm{s}$. Next, we use our proposed Natias method to calculate neuron attributions in the target layer along the straight line $\bm{s}'+\gamma(\bm{s}-\bm{s}')$ and compute the gradient map $\bm{G}_t$ to corrupt critical features in each iteration $t$:
\begin{equation}
    \bm{G}_t=\frac{\partial L(\bm{s}^t_{adv})}{\partial \bm{s}^t_{adv}},
\end{equation}
where $\bm{s}'$ is the baseline image and $\bm{s}^t_{adv}$ represents the modified stego image in the current iteration. Then, we leverage a mask $\bm{m}$ to control whether elements can
be enhanced, and $\bm{m}=\bm{0}$ in the initial state. If the element was modified before, the corresponding flag in $\bm{m}$ will be set to $1$. Let $\bm{G}'_t=\bm{G}_t\cdot(\bm{1}-\bm{m})$, select the top $k$ elements from $\bm{G}'_t$ and set the corresponding positions in $\bm{m}$ to $1$. We construct the enhanced cover $\bm{e}^{t+1}=\bm{e}^t+\bm{G}'_t$ and obtain the corresponding stego $\bm{s}^{t+1}$, where $\bm{e}^{t+1}$ is the enhanced cover in the iteration $t$ and $\bm{e}^0=\bm{c}$. When the probabilistic output of $\bm{s}^t$ is larger than a threshold $\tau$, we scramble the secret message and regenerate the stego image. Repeat the above process until the stego corresponding to the enhanced cover can deceive the target steganalytic model. Otherwise, the conventional stego is used as the final stego.

\begin{algorithm}[H]
\renewcommand{\algorithmicrequire}{\textbf{INPUT:}}
\renewcommand{\algorithmicensure}{\textbf{OUTPUT:}}
\caption{Natias combined with ADV-EMB scheme}\label{alg}
\begin{algorithmic}[1]
\REQUIRE {target steganalytic model $\mathscr{F}$, target layer $\bm{y}$, initial distortion $\bm{\rho}=\{\bm{\rho}^+,\bm{\rho}^-\}$, cover image $
\bm{c}$, secret message $\bm{msg}$, initial parameter $\beta$, baseline image $\bm{x}'$, sample step number $M$}
\ENSURE {adversarial stego image $\bm{s}$}
\STATE $\beta\leftarrow0$, $M\leftarrow30$, $\bm{z}\leftarrow\bm{c}$, $A_{\bm{y}}\leftarrow\bm{0}$
\WHILE {$\beta\leq1$} 
\STATE Randomly select $1-\beta$ of the pixels from $\bm{c}$ as $\bm{c}_{com}$
\STATE The remaining pixels are denoted as $\bm{c}_{adj}$
\STATE Select the first $1-\beta$ of the bits from $\bm{msg}$ as $\bm{msg}_1$
\STATE The remaining bits are denoted as $\bm{msg}_2$
\STATE $\bm{z} \leftarrow \bm{Emb}(\bm{c}_{com}, \bm{msg}_1, \bm{\rho})\cup{\bm{c}_{adj}}$

\STATE $\bm{z}' \leftarrow \bm{Emb}(\bm{c}_{com}, \bm{msg}_1, \bm{\rho})\cup(\bm{c}_{adj}-\bm{1})$
\FOR {$m\leftarrow1\ to\ M$}
\STATE $A_{\bm{y}}\leftarrow A_{\bm{y}}+\nabla_{\bm{y}(\bm{z}'+\frac{m}{M}(\bm{z}-\bm{z}'))}{\mathscr{F}(\bm{z}'+\frac{m}{M}(\bm{z}-\bm{z}'))}$
\ENDFOR
\STATE $A_{\bm{y}}\leftarrow \frac{1}{M}(\bm{y}-\bm{y}')\cdot A_{\bm{y}}$
\STATE Compute the gradient map $\bm{G}_{\bm{z}}$ by Eq.~(\ref{adv_emb_grad})
\STATE Compute $\bm{\rho}_{attr}$ according to Eq.~(\ref{adjust})
\STATE Update $\bm{z}$ according to Eq.~(\ref{disjoint})
\IF {$\mathscr{F}(\bm{z})=0$} 
\STATE $\bm{s}\leftarrow\bm{z}$ 
\ENDIF
\STATE $\beta\leftarrow\beta+0.1$
\ENDWHILE
\STATE $\bm{s}\leftarrow \bm{Emb}(\bm{c}, \bm{msg}, \bm{\rho})$
\end{algorithmic}
\label{alg1}
\end{algorithm}

The main process of our proposed Natias combined with distortion adjustment based methods like ADV-EMB is elucidated as follows. Given a cover image $\bm{c}$ with $n$ pixels and a secret message $\bm{msg}$, we first calculate the initial distortion $\bm{\rho}=\{\bm{\rho}^+,\bm{\rho}^-\}$ by using existing distortion 
functions and initialize a parameter $\beta=0$. Continuing, we randomly divide the cover image into two non-overlapping groups: a common group containing $1-\beta$ of the cover pixels and an adjustable group containing the remaining cover pixels. Subsequently, we embed $\bm{msg}_1$ including $1-\beta$ of the secret message bits into the common group based on the initial distortion using the embedding simulator~\cite{min_distortion_model} as follows:
\begin{equation}
    \bm{z} = \bm{Emb}(\bm{c}_{com}, \bm{msg}_1, \bm{\rho})\cup{\bm{c}_{adj}},
\end{equation}
where $\bm{c}_{com}$ and $\bm{c}_{adj}$ are the common group and adjustable group of the cover image, respectively. In this case, we subtract one from all pixels in the adjustable group of the cover image after conventional embedding to zero as the baseline image $\bm{z}'$, i.e., $\bm{z}'=\bm{Emb}(\bm{c}_{com}, \bm{msg}_1, \bm{\rho})\cup(\bm{c}_{adj}-\bm{1})$. We input $\bm{z}$ and the baseline image $\bm{z}'$ into the target steganalytic model to obtain the logits of prediction results and the neuron output of the target layer. Then we determine the contribution of each neuron to the steganalytic model's decision based on neuron attribution as in Eq.~(\ref{riemann}). It is worth noting that here we choose the straight line $\bm{z}'+\gamma(\bm{z}-\bm{z}')$ for calculating the path integral.
Following this, we solve the minimization problem presented in Eq.~(\ref{optimize}) and obtain the gradient map:
\begin{equation}\label{adv_emb_grad}
    \bm{G}_{\bm{z}}=\{g_{\bm{z}1}, g_{\bm{z}2}, ..., g_{\bm{z}n}\}=\frac{\partial L(\bm{z})}{\partial \bm{z}}.
\end{equation}
Next, we adjust the initial distortion to attribution distortion $\bm{\rho}_{attr}=\{\bm{\rho}_{attr}^+,\bm{\rho}_{attr}^-\}$ based on the obtained gradient map as follows:
\begin{equation}\label{adjust}
    \begin{array}{l}
    \rho_{attr+}^{i}=\left\{\begin{array}{ll}
    \rho_i^+ / \alpha, & g_{\bm{z}i}<0 \\
    \rho_i^+, & g_{\bm{z}i}=0 \\
    \rho_i^+ * \alpha, & g_{\bm{z}i}>0
    \end{array}\right. \\
    \rho_{attr-}^{i}=\left\{\begin{array}{ll}
    \rho_i^- / \alpha, & g_{\bm{z}i}>0 \\
    \rho_i^-, & g_{\bm{z}i}=0 \\
    \rho_i^- * \alpha, & g_{\bm{z}i}<0
    \end{array}\right.
    \end{array},
\end{equation}
where $\alpha$ is the scaling factor larger than $1$. Finally, based on the adjusted distortion, we embed $\bm{msg}_2$ with the remaining secret message bits into the adjustable group:
\begin{equation}\label{disjoint}
    \bm{z} = \bm{Emb}(\bm{c}_{com}, \bm{msg}_1, \bm{\rho})\cup{\bm{Emb}(\bm{c}_{adj}, \bm{msg}_2, \bm{\rho}_{attr})}.
\end{equation}
In this way, we have embedded all secret message bits into the cover image. If $\bm{z}$ can deceive the target steganalytic model, i.e., $\mathscr{F}(\bm{z})=0$, then we use $\bm{z}$ as the final stego image $\bm{s}$. Otherwise, we update $\beta=\beta+0.1$ and repeat the above steps until $\bm{z}$ can deceive the target steganalytic model. The whole process of our proposed Natias combined with ADV-EMB scheme is illustrated in Algorithm~\ref{alg}.

In addition, our proposed method can be integrated with stego post-processing based methods such as USGS~\cite{stego_selection}. First, we input the cover image into the pre-trained target steganalytic model to obtain neuron attribution values. Then, based on the attribution results, we corrupt critical features and obtain the gradient map from backpropagation. Subsequently, we adjust the distortion in regions with significant gradients and minimized distortions by Eq.~(\ref{adjust}). After repeating this process multiple times, we could obtain $N$ different distortion maps. Next, based on these distortion maps, we embed the exact same secret message into the cover to obtain $N+1$ candidate stego images, including the initial stego image. Then, these stego images capable of deceiving the target steganalytic model are selected. Finally, we obtain the set $H_{\bm{c}}$ of adaptive high-pass filters for the given cover $\bm{c}$ and employ the resulting set $H_{\bm{c}}$ to calculate the corresponding image residuals of the cover image and those available candidate stego images, where the construction of $H_{\bm{c}}$ follows the settings in the paper~\cite{stego_selection}. The image with the minimal residual distance between the cover is chosen as the ultimate stego image.

\section{Experiments}\label{experiments}

\subsection{Experimental Settings}
The experiments in this paper are conducted on $20,000$ grayscale images, which consist of two widely studied datasets in the steganography field, BOSSBase ver.1.01~\cite{BOSS} and BOWS2~\cite{BOWS2}. Each contains $10,000$ grayscale images of $512\times512$. To match the settings of previous works, the original images are resized to $256\times256$ by $\textit{imresize}()$ of $\text{MatLab}$ with the default settings. For the CNN-based steganalytic models, we randomly divide this dataset into three non-overlapping parts, each containing $14,000$, $1,000$, and $5,000$ images, designated as the training set, the validation set, and the test set, respectively. For traditional feature-based steganalytic models, $10,000$ images are randomly chosen for training, and the rest $10,000$ images are for testing. Two basic steganographic distortion functions including HILL and SUNIWARD, three adversarial steganographic methods including cover enhancement based method SPS-ENH, distortion adjustment based method ADV-EMB, and stego post-processing based method USGS, four CNN-based steganalytic models (SRNet, CovNet, LWENet, and SiaStegNet), and one traditional feature-based steganalytic model SRM are included to evaluate the effectiveness of our proposed method. We use Natias-SPS to denote the version of Natias combined with SPS-ENH, Natias-ADV to denote the version of Natias combined with ADV-EMB, and Natias-USGS to denote the version of Natias combined with USGS.

\begin{table}
\renewcommand{\arraystretch}{0.2}
\centering
\caption{Detection accuracy ($\%$) of the target pre-trained steganalytic model compared with SPS-ENH. The target steganalytic model is CovNet.}
\label{ASR-SPS-ENH}
\renewcommand{\arraystretch}{1.5}
\begin{tabular}{ccccc}
\toprule
Payload & \multicolumn{2}{c}{$0.2$ bpp} & \multicolumn{2}{c}{$0.4$ bpp}                                    \\ 
\cline{1-5}
Mehod & SPS-ENH & Natias-SPS & SPS-ENH & Natias-SPS \\ 
\midrule
HILL & $36.48$ & $36.48$ & $42.28$ & $42.28$   \\
SUNIWARD & $40.70$ & $40.70$ & $45.01$ & $44.97$   \\
\bottomrule
\end{tabular}
\end{table}

The training details of the CNN-based steganalytic models (SRNet, CovNet, LWENet, and SiaStegNet) are the same as reported in ~\cite{SRNet}-~\cite{LWENet}, including the batch size, learning rate, weight decay, and the training epochs. CovNet, LWENet, and SiaStegNet are trained from scratch directly for all payloads, while SRNet is trained with the first curriculum training method as reported in ~\cite{SRNet}. In other words, the detectors for payloads of $0.1$, $0.2$, and $0.3$ bpp (bit per pixel) were trained by seeding with a network trained for payload $0.4$ bpp. In addition, we repeat the experiment ten times and take the mean value of the ten results as the final result to increase credibility when using SRM as the steganalyzer. Unless otherwise specified, in this paper we employ the embedding simulator algorithm for message embedding.

Regarding the issue of target layer selection, when attacking SRNet, we treat its type $3$ layers as a whole layer and select it as the target layer. Similarly, when attacking CovNet, we select its group $3$ as the target layer, and when attacking LWENet, we select its layer $8$ as the target layer. We will follow the same configuration for subsequent experiments. For a fair comparison, we adhere to the settings of corresponding papers for various relevant hyperparameters. We set the sample step number $M=50$ to calculate the attribution results and the median of the absolute values of all attribution results as the threshold $T$ to select important neurons. The weight parameter $\lambda$ in Eq.~(\ref{loss}) is set as $1$.

\subsection{Performance against the Target Steganalytic Model}\label{asr}
In this section, we compare the performance of different adversarial steganographic methods against the target steganalytic model. We designate SRNet, CovNet, and LWENet as the target steganalytic models and employ SPS-ENH, ADV-EMB, USGS, and our proposed Natias to generate stego images to deceive them. The target steganalytic models are all adversary-unaware, i.e., we train them on the stego images obtained from basic distortion functions (HILL and SUNIWARD) in corresponding payload cases.

The detection accuracy results of the target steganalytic model are reported in Table~\ref{ASR-SPS-ENH}, Table~\ref{ASR-ADV-EMB}, and Table~\ref{ASR-USGS} for comparison with SPS-ENH, ADV-EMB, and USGS. It can be observed that our method essentially achieves comparable performance with initial methods. When the payload is higher, our method can outperform them. This is because in lower payload cases, the amount of steganographic modification is relatively small, making it difficult to sufficiently corrupt the critical features. But in higher payload cases, we can thoroughly corrupt them and effectively deceive the target steganalytic model. 

Due to space constraints and the prevalent utilization of distortion adjustment based methods in adversarial steganography, when compared with SPS-ENH and USGS, we conduct experiments only in $0.2$ bpp and $0.4$ bpp cases and select CovNet as the target steganalytic model. When compared with ADV-EMB, we conduct experiments in all payload cases and select SRNet, CovNet, and LWENet as the target steganalytic models. Unless otherwise specified, experiments involving them will be conducted with the same settings.

\begin{table*}
\centering
\caption{Detection accuracy ($\%$) of the target pre-trained steganalytic model compared with ADV-EMB.}
\label{ASR-ADV-EMB}
\renewcommand{\arraystretch}{1.5}
\begin{tabular}{cccccccccc}
\toprule
\multirow{2}{*}{Target Model}                   & Payload & \multicolumn{2}{c}{$0.1$ bpp}  & \multicolumn{2}{c}{$0.2$ bpp}   & \multicolumn{2}{c}{$0.3$ bpp}   & \multicolumn{2}{c}{$0.4$ bpp}                                    \\ 
\cline{2-10}
& Mehod & ADV-EMB & Natias-ADV & ADV-EMB & Natias-ADV & ADV-EMB & Natias-ADV & ADV-EMB & Natias-ADV\\ 
\midrule
\multirow{2}{*}{SRNet} & HILL & $\bm{38.93}$ & $38.96$ & $\bm{38.80}$ & $38.95$ & $40.75$ &$\bm{40.72}$ & $42.68$ & $\bm{42.50}$  \\
& SUNIWARD & $\bm{41.55}$ & $41.78$ & $44.90$ & $\bm{44.36}$ & $45.55$ &$\bm{45.21}$ & $48.44$ & $\bm{48.03}$  \\
\midrule
\multirow{2}{*}{CovNet} & HILL & $36.58$ & $\bm{36.30}$ & $38.75$ & $\bm{38.54}$ & $40.15$ &$\bm{39.94}$ & $42.66$ & $\bm{42.49}$  \\
& SUNIWARD & $41.51$ & $\bm{41.48}$ & $45.85$ & $\bm{44.87}$ & $47.18$ &$\bm{45.51}$ & $48.48$ & $\bm{46.85}$  \\
\midrule
\multirow{2}{*}{LWENet} & HILL & $35.04$ & $\bm{34.82}$ & $40.27$ & $\bm{39.89}$ & $44.05$ &$\bm{43.42}$ & $43.72$ & $\bm{43.33}$  \\
& SUNIWARD & $42.66$ & $\bm{42.31}$ & $45.56$ & $\bm{43.93}$ & $48.68$ &$\bm{46.80}$ & $50.71$ & $\bm{48.61}$  \\                      
\bottomrule
\end{tabular}
\end{table*}

\begin{table}
\centering
\caption{Detection accuracy ($\%$) of the target pre-trained steganalytic model compared with USGS. The target steganalytic model is CovNet.}
\label{ASR-USGS}
\renewcommand{\arraystretch}{1.5}
\begin{tabular}{ccccc}
\toprule
Payload & \multicolumn{2}{c}{$0.2$ bpp} & \multicolumn{2}{c}{$0.4$ bpp}                                    \\ 
\cline{1-5}
Mehod & USGS & Natias-USGS & USGS & Natias-USGS \\ 
\midrule
HILL & $39.67$ & $\bm{39.40}$ & $43.71$ & $\bm{43.01}$   \\
SUNIWARD & $44.44$ & $\bm{43.88}$ & $48.75$ & $\bm{47.31}$   \\
\bottomrule
\end{tabular}
\end{table}

\subsection{Performance against Non-target Steganalytic models}\label{transferability}
In this section, we evaluate the transferability of our proposed method, i.e., the ability of adversarial steganography to deceive non-target steganalytic models. We quantify this using the detection accuracy of non-target steganalytic models when steganalyzing stego images generated by attacking the target steganalytic model. The lower the detection accuracy of the non-target classifier is, the better the transferability of adversarial steganography. Additionally, to demonstrate the superiority of our proposed Natias compared with the initial methods, we exhibit the average improvement in resisting different non-target steganalytic models.

We employ four adversarial steganographic methods to attack the target steganalytic model and then use different non-target steganalytic models for steganalysis. The experimental results compared with SPS-ENH are exhibited in Table~\ref{trans-SPS-ENH-cov}, and the target steganalytic model is CovNet. The experimental results compared with ADV-EMB are exhibited in Table~\ref{trans-ADV}, and SRNet, CovNet, and LWENet are selected as target steganalytic models. And the experimental results compared with USGS are shown in Table~\ref{trans-USGS-cov}, and the target steganalytic model is CovNet. Those values in bolded text denote the best results in the corresponding cases, and the ``average" column in these tables denotes the average improvements over four steganalytic models for a given steganographic algorithm and payload.

It can be observed that our proposed Natias achieves almost the best performance in all testing scenarios. As shown in Table~\ref{trans-ADV}, in lower payload cases, the transferability of Natias-ADV is comparable to ADV-EMB, and there is a slight average improvement over four different steganalytic models. Additionally, for a given target steganalytic model and basic distortion function, as the embedding rate increases, the average improvement magnitude of transferability performance also increases. We infer that in lower payload cases, the amount of steganographic modification is relatively small, making it difficult to sufficiently corrupt the critical features. Thus, this leads to a restricted enhancement of transferability. Conversely, in higher payload cases, we can thoroughly corrupt them and effectively deceive the target model. 

In addition, as illustrated in Table~\ref{trans-ADV}, when the target steganalytic model is SRNet, the transferability of Natias-ADV is comparable to ADV-EMB. However, when the target model is CovNet or LWENet, Natias-ADV exhibits a noticeable improvement compared with ADV-EMB on transferability. We infer that SRNet is the weakest among all the CNN-based steganalytic models involved, so the intermediate layer features it relies on for classification are less crucial. Consequently, it is challenging for the generated stego images to deceive other steganalytic models with stronger performance. The experimental results compared with SPS-ENH and USGS are exhibited in Table~\ref{trans-SPS-ENH-cov} and Table~\ref{trans-USGS-cov}, respectively. These results indicate that our proposed Natias completely outperforms SPS-ENH and USGS against different non-target steganalytic models. 

Furthermore, we present some visual experimental results to illustrate the effectiveness of our method. We employ the attribution analysis method called integrated gradients to generate attention distributions for steganalytic models. Specifically, we first randomly select a cover image from the dataset and input it into three different steganalytic models, including SRNet, CovNet, and LWENet, to obtain the corresponding attention distributions. Then, we use CovNet as the target steganalytic model and employ Natias-ADV to generate the corresponding stego image. Next, we input the stego into the aforementioned steganalytic models to obtain new attention distributions. The attention distributions of the three steganalytic models when detecting the cover and stego images are shown in Fig.~\ref{attention_map}. It can be observed that when detecting the cover, the steganalytic model's attention is mainly concentrated on the texture regions. However, when detecting the stego, due to the corruption of critical features, part of the steganalytic model's attention shifts to the smooth region. Though we select CovNet as the target model, similar phenomena can also be observed when the stego is input into SRNet and LWENet. This enables our method to successfully deceive non-target steganalytic models.

According to the experimental results presented above, our method indeed significantly enhance the transferability of adversarial steganography. Even when selecting the relatively weaker SRNet as the target steganalytic model, there is an $2.39\%$ average improvement compared with ADV-EMB in the $0.4$ bpp case when the basic distortion function is HILL.

\begin{table*}
\centering
\caption{Detection accuracy ($\%$) of different non-target steganalytic models when detecting our proposed Natias and SPS-ENH. The target steganalytic model is CovNet. Those values in bolded text denote the best results in the corresponding cases. The ``Average" column denotes the gain in security measured using the best steganalyzer amongst SPS-ENH and Natias.}
\label{trans-SPS-ENH-cov}
\renewcommand{\arraystretch}{1.5}
\begin{tabular}{ccccccccccc}
\toprule
\multirow{2}{*}{Method}   & \multirow{2}{*}{Payload}  & \multicolumn{2}{c}{SRNet}  & \multicolumn{2}{c}{SiaStegNet}   & \multicolumn{2}{c}{LWENet}   & \multicolumn{2}{c}{SRM} & \multirow{2}{*}{Average} \\ 
\cline{3-10}
&   & SPS-ENH & Natias-SPS & SPS-ENH & Natias-SPS & SPS-ENH & Natias-SPS & SPS-ENH & Natias-SPS\\ 
\midrule
\multirow{2}{*}{HILL}
& $0.2$ bpp & $57.02$ & $\bm{54.60}$ & $57.17$ & $\bm{54.87}$ & $56.61$ &$\bm{53.97}$ & $58.23$ & $\bm{57.33}$ & $\downarrow2.07$ \\
& $0.4$ bpp & $70.59$ & $\bm{69.14}$ & $72.73$ & $\bm{69.68}$ & $71.60$ &$\bm{68.83}$ & $70.22$ & $\bm{69.23}$ & $\downarrow2.07$ \\
\midrule
\multirow{2}{*}{SUNIWARD} 
& $0.2$ bpp & $63.16$ & $\bm{59.84}$ & $61.70$ & $\bm{57.92}$ & $61.80$ &$\bm{59.58}$ & $62.27$ & $\bm{61.62}$ & $\downarrow2.49$ \\
& $0.4$ bpp & $69.71$ & $\bm{67.28}$ & $68.07$ & $\bm{66.41}$ & $67.18$ &$\bm{66.01}$ & $74.90$ & $\bm{73.94}$ & $\downarrow1.56$  \\           
\bottomrule
\end{tabular}
\end{table*}

\begin{table*}
\centering
\caption{Detection accuracy ($\%$) of different non-target steganalytic models when detecting our proposed Natias and ADV-EMB. Those values in bolded text denote the best results in the corresponding cases. The ``Average" column denotes the gain in security measured using the best steganalyzer amongst ADV-EMB and Natias.}
\label{trans-ADV}
\renewcommand{\arraystretch}{1.5}
\setlength{\tabcolsep}{3pt}
\begin{tabularx}{\textwidth}{ccccccccccccc}
\toprule
\multirow{2}{*}{Target Model} &  \multirow{2}{*}{Method}   & \multirow{2}{*}{Payload}  & ADV-EMB & Natias-ADV & ADV-EMB & Natias-ADV & ADV-EMB & Natias-ADV & ADV-EMB & Natias-ADV & \multirow{2}{*}{Average}\\ 
\cline{4-11}
&  &   & \multicolumn{2}{c}{CovNet}  & \multicolumn{2}{c}{SiaStegNet}   & \multicolumn{2}{c}{LWENet}   & \multicolumn{2}{c}{SRM} &  \\ 

\midrule
\multirow{8}{*}{SRNet} & \multirow{4}{*}{HILL} & $0.1$ bpp & $57.90$ & $\bm{57.57}$ & $54.87$ & $\bm{54.56}$ & $57.02$ &$\bm{56.08}$ & $52.51$ & $\bm{52.46}$ & $\downarrow0.41$  \\
& & $0.2$ bpp & $64.31$ & $\bm{63.26}$ & $58.35$ & $\bm{58.26}$ & $61.21$ &$\bm{60.63}$ & $55.17$ & $\bm{54.66}$ & $\downarrow0.56$ \\
& & $0.3$ bpp & $70.42$ & $\bm{69.37}$ & $61.59$ & $\bm{60.84}$ & $63.64$ &$\bm{62.47}$ & $57.45$ & $\bm{56.67}$ & $\downarrow0.95$ \\
& & $0.4$ bpp & $69.35$ & $\bm{67.48}$ & $68.38$ & $\bm{66.83}$ & $68.95$ &$\bm{64.58}$ & $63.96$ & $\bm{62.18}$ & $\downarrow2.39$ \\
\cmidrule(lr){2-12}
 &\multirow{4}{*}{SUNIWARD} & $0.1$ bpp & $61.88$ & $\bm{61.49}$ & $60.44$ & $\bm{60.23}$ & $61.49$ &$\bm{61.34}$ & $55.86$ & $\bm{55.59}$ & $\downarrow0.26$ \\
& & $0.2$ bpp & $72.00$ & $\bm{71.26}$ & $69.25$ & $\bm{68.91}$ & $71.99$ &$\bm{71.01}$ & $62.34$ & $\bm{61.30}$ & $\downarrow0.78$ \\
& & $0.3$ bpp & $79.99$ & $\bm{79.49}$ & $76.98$ & $\bm{76.15}$ & $79.24$ &$\bm{78.04}$ & $66.77$ & $\bm{64.90}$ & $\downarrow1.10$ \\
& & $0.4$ bpp & $83.28$ & $\bm{81.43}$ & $79.56$ & $\bm{78.12}$ & $83.35$ &$\bm{82.15}$ & $69.51$ & $\bm{66.64}$ & $\downarrow1.84$  \\ 
\midrule
&  &  & \multicolumn{2}{c}{SRNet}  & \multicolumn{2}{c}{SiaStegNet}   & \multicolumn{2}{c}{LWENet}   & \multicolumn{2}{c}{SRM}  & \\ 
\midrule
\multirow{8}{*}{CovNet} & \multirow{4}{*}{HILL} & $0.1$ bpp & $52.91$ & $\bm{51.40}$ & $54.18$ & $\bm{53.55}$ & $54.60$ &$\bm{53.83}$ & $52.48$ & $\bm{51.92}$ & $\downarrow0.87$ \\
& & $0.2$ bpp & $56.92$ & $\bm{54.26}$ & $58.16$ & $\bm{57.64}$ & $58.41$ &$\bm{56.34}$ & $55.63$ & $\bm{54.25}$ & $\downarrow1.66$ \\
& & $0.3$ bpp & $59.23$ & $\bm{56.11}$ & $60.23$ & $\bm{59.29}$ & $58.99$ &$\bm{56.93}$ & $57.03$ & $\bm{55.51}$ & $\downarrow2.66$ \\
& & $0.4$ bpp & $64.13$ & $\bm{60.39}$ & $67.59$ & $\bm{64.17}$ & $66.11$ &$\bm{62.64}$ & $63.96$ & $\bm{60.72}$ & $\downarrow3.47$ \\
\cmidrule(lr){2-12}
& \multirow{4}{*}{SUNIWARD} & $0.1$ bpp & $59.19$ & $\bm{58.36}$ & $59.24$ & $\bm{58.87}$ & $59.49$ &$\bm{59.06}$ & $56.17$ & $\bm{56.13}$ & $\downarrow0.42$ \\
& & $0.2$ bpp & $65.17$ & $\bm{63.00}$ & $69.25$ & $\bm{65.02}$ & $68.36$ &$\bm{66.28}$ & $62.86$ & $\bm{61.98}$ & $\downarrow2.34$ \\
& & $0.3$ bpp & $70.31$ & $\bm{67.47}$ & $74.04$ & $\bm{71.12}$ & $73.78$ &$\bm{70.72}$ & $65.99$ & $\bm{63.60}$ & $\downarrow2.80$ \\
& & $0.4$ bpp & $76.41$ & $\bm{72.94}$ & $77.11$ & $\bm{74.33}$ & $79.40$ &$\bm{76.58}$ & $69.27$ & $\bm{65.27}$  & $\downarrow3.27$ \\
\midrule
&  &  & \multicolumn{2}{c}{SRNet}  & \multicolumn{2}{c}{CovNet}   & \multicolumn{2}{c}{SiaStegNet}   & \multicolumn{2}{c}{SRM}  & \\ 
\midrule
\multirow{8}{*}{LWENet} & \multirow{4}{*}{HILL} & $0.1$ bpp & $53.68$ & $\bm{52.41}$ & $56.51$ & $\bm{54.85}$ & $53.27$ &$\bm{52.70}$ & $52.64$ & $\bm{52.30}$ & $\downarrow0.96$  \\
& & $0.2$ bpp & $58.89$ & $\bm{56.90}$ & $63.33$ & $\bm{61.50}$ & $57.71$ &$\bm{56.76}$ & $55.97$ & $\bm{55.05}$ & $\downarrow1.42$ \\
& & $0.3$ bpp & $64.08$ & $\bm{60.87}$ & $69.78$ & $\bm{68.45}$ & $61.56$ &$\bm{60.49}$ & $59.16$ & $\bm{57.79}$ & $\downarrow1.75$ \\
& & $0.4$ bpp & $62.51$ & $\bm{59.71}$ & $63.46$ & $\bm{61.04}$ & $65.64$ &$\bm{63.40}$ & $64.42$ & $\bm{62.41}$ & $\downarrow2.37$ \\
\cmidrule(lr){2-12}
& \multirow{4}{*}{SUNIWARD} & $0.1$ bpp & $58.96$ & $\bm{58.10}$ & $59.64$ & $\bm{58.75}$ & $58.54$ &$\bm{57.96}$ & $56.35$ & $\bm{56.28}$ & $\downarrow0.60$  \\
& & $0.2$ bpp & $65.11$ & $\bm{63.08}$ & $68.53$ & $\bm{66.84}$ & $66.49$ &$\bm{64.68}$ & $63.09$ & $\bm{62.34}$ & $\downarrow1.57$ \\
& & $0.3$ bpp & $70.37$ & $\bm{67.21}$ & $75.59$ & $\bm{72.10}$ & $73.67$ &$\bm{70.95}$ & $66.91$ & $\bm{64.46}$ & $\downarrow2.96$ \\
& & $0.4$ bpp & $75.76$ & $\bm{71.53}$ & $78.32$ & $\bm{74.41}$ & $76.77$ &$\bm{71.96}$ & $69.92$ & $\bm{67.10}$ & $\downarrow3.94$ \\
\bottomrule
\end{tabularx}
\end{table*}

\begin{table*}
\centering
\caption{Detection accuracy ($\%$) of different non-target steganalytic models when detecting our proposed Natias and USGS. The target steganalytic model is CovNet. Those values in bolded text denote the best results in the corresponding cases. The ``Average" column denotes the gain in security measured using the best steganalyzer amongst USGS and Natias.}
\label{trans-USGS-cov}
\renewcommand{\arraystretch}{1.5}
\begin{tabular}{ccccccccccc}
\toprule
\multirow{2}{*}{Method}   & \multirow{2}{*}{Payload}  & \multicolumn{2}{c}{SRNet}  & \multicolumn{2}{c}{SiaStegNet}   & \multicolumn{2}{c}{LWENet}   & \multicolumn{2}{c}{SRM} & \multirow{2}{*}{Average} \\ 
\cline{3-10}
&   & USGS & Natias-USGS & USGS & Natias-USGS & USGS & Natias-USGS & USGS & Natias-USGS\\ 
\midrule
\multirow{2}{*}{HILL}
& $0.2$ bpp & $51.93$ & $\bm{50.65}$ & $\bm{52.18}$ & $53.51$ & $54.51$ &$\bm{51.15}$ & $54.09$ & $\bm{53.63}$ & $\downarrow0.94$ \\
& $0.4$ bpp & $60.55$ & $\bm{58.05}$ & $62.04$ & $\bm{59.89}$ & $59.15$ &$\bm{57.15}$ & $64.14$ & $\bm{63.23}$ & $\downarrow1.89$ \\
\midrule
\multirow{2}{*}{SUNIWARD} 
& $0.2$ bpp & $58.82$ & $\bm{57.14}$ & $59.33$ & $\bm{58.86}$ & $59.35$ &$\bm{58.12}$ & $60.49$ & $\bm{59.83}$ & $\downarrow1.01$ \\
& $0.4$ bpp & $69.35$ & $\bm{67.86}$ & $70.17$ & $\bm{68.43}$ & $70.10$ &$\bm{68.00}$ & $68.92$ & $\bm{67.42}$ & $\downarrow1.71$  \\           
\bottomrule
\end{tabular}
\end{table*}

\subsection{Performance against Retrained Steganalytic models}
As described in the previous Sec.~\ref{transferability}, our proposed method can effectively enhance transferability to deceive non-target steganalytic models. However, when the adversarial steganographic method we use is exposed, adversaries can retrain the targeted steganalytic model based on the generated adversarial stego images or employ other non-target steganalytic models to improve detection capabilities. In this section, we evaluate the performance of our proposed method in resisting detection by three retrained target steganalytic models and two other non-target steganalytic models. 

\begin{figure*}[!t]
\centering
\subfloat[SRNet]{\includegraphics[width=0.3\linewidth]{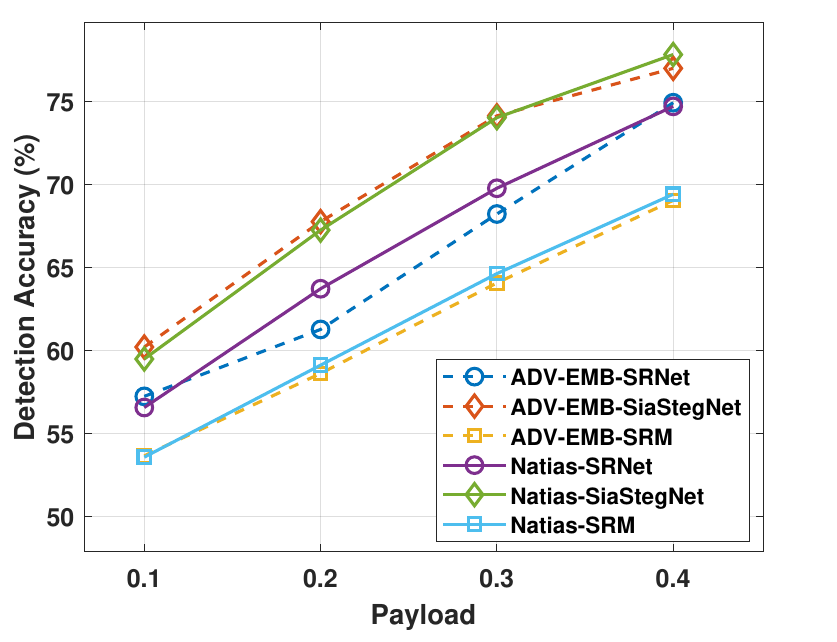}}
\hfil
\subfloat[CovNet]{\includegraphics[width=0.3\linewidth]{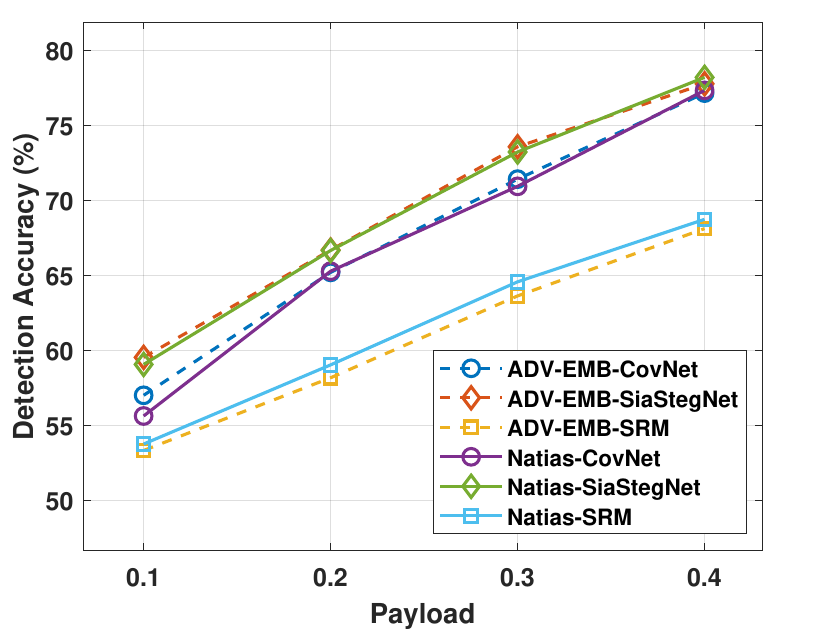}}
\hfil
\subfloat[LWENet]{\includegraphics[width=0.3\linewidth]{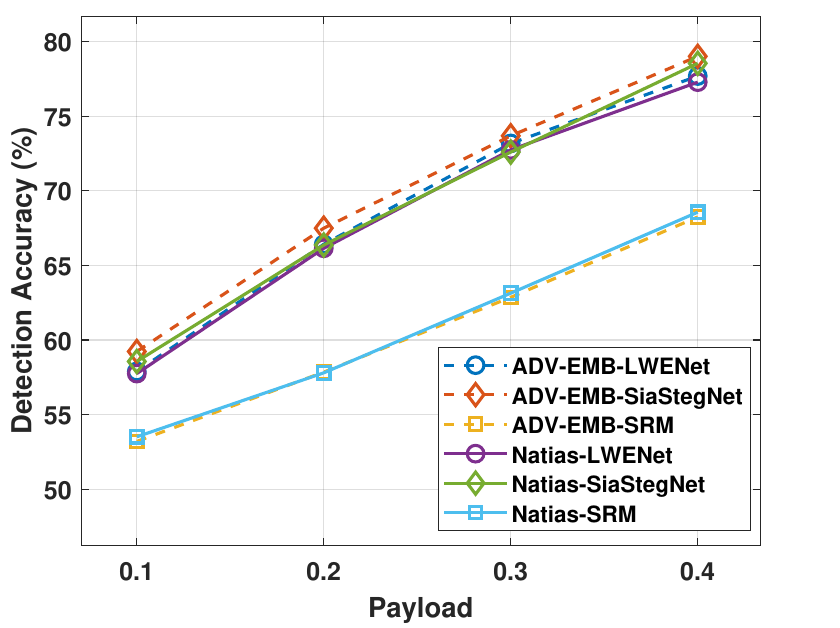}}\\
\subfloat[SRNet]{\includegraphics[width=0.3\linewidth]{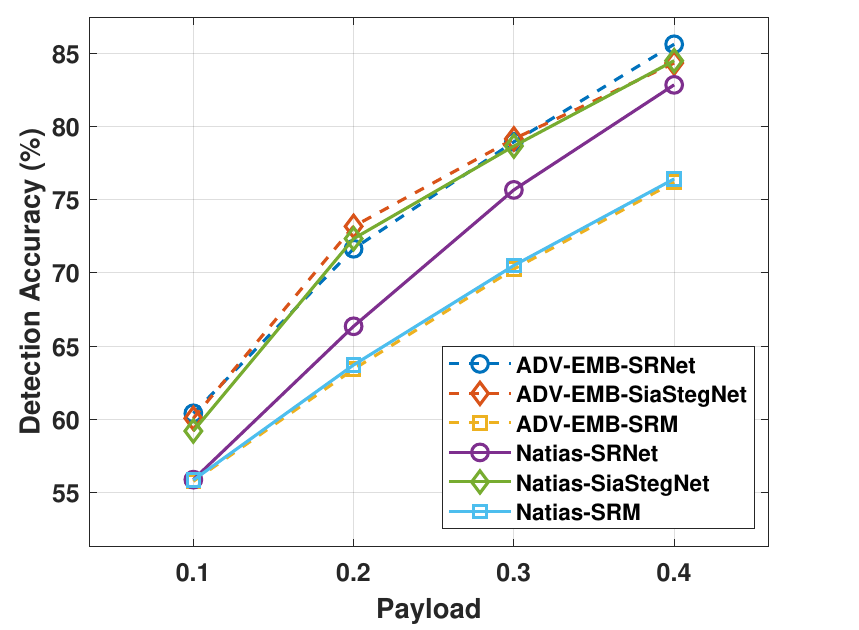}}
\hfil
\subfloat[CovNet]{\includegraphics[width=0.3\linewidth]{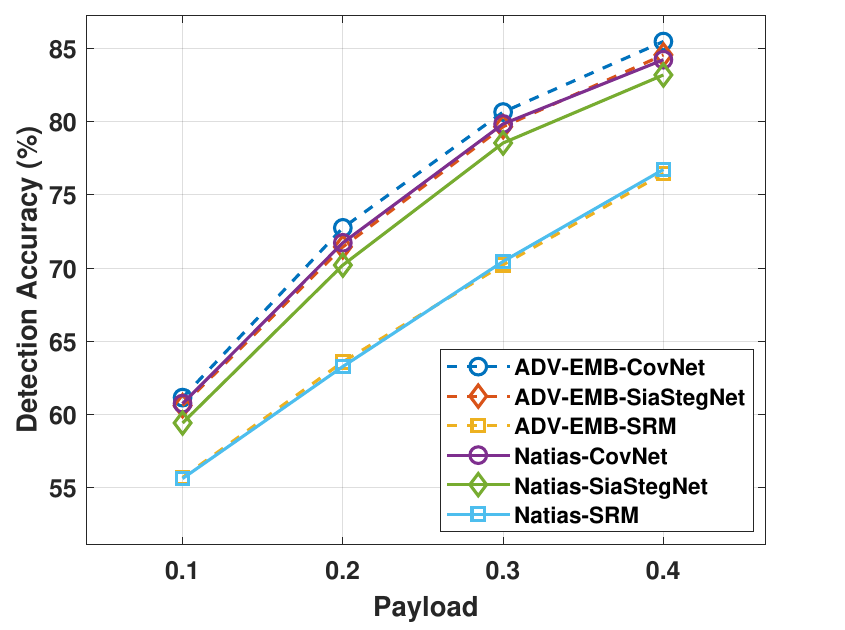}}
\hfil
\subfloat[LWENet]{\includegraphics[width=0.3\linewidth]{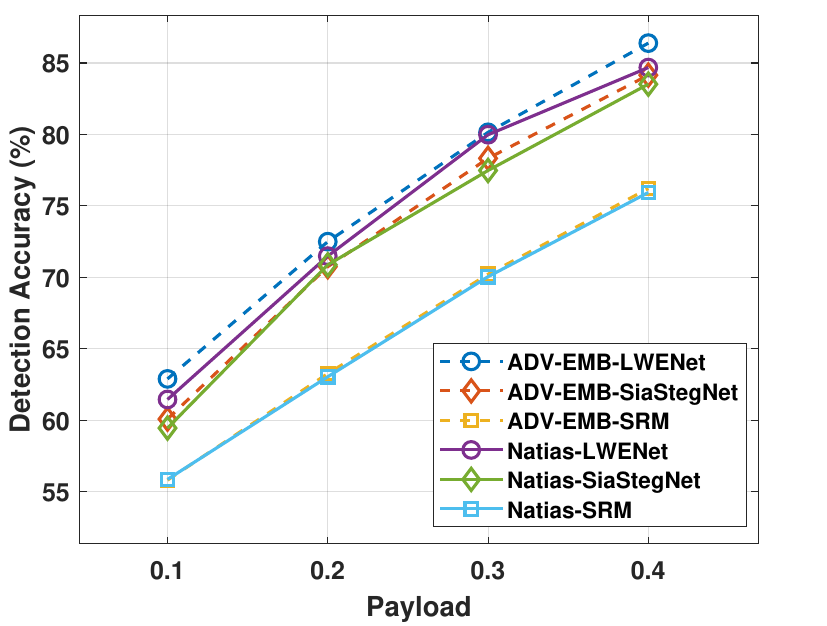}}
\caption{Detection accuracy evaluated on retrained steganalytic models compared with ADV-EMB. The y-axis represents the detection accuracy of the steganalytic model, and the x-axis represents the payload. The basic distortion function of the top row is HILL, and the basic distortion function of the bottom row is SUNIWARD.}
\label{retrain}
\end{figure*}

We first use SRNet, CovNet, and LWENet as target steganalytic models, and then employ SPS-ENH, USGS, ADV-EMB, and Natias to generate stego images. Subsequently, we retrain each steganalytic model based on the corresponding stego images. In addition, we employ a CNN-based steganalytic model SiaStegNet and a traditional feature-based steganalytic model SRM to detect the generated stego images. The experimental results of the detection accuracy of retrained target steganalytic models and other non-target steganalytic models are illustrated in Table~\ref{retrain-SPS-ENH-cov}, Fig.~\ref{retrain}, and Table~\ref{retrain-USGS-cov}. In Fig.~\ref{retrain}, ``ADV-EMB-SRNet" represents the target steganalytic model SRNet obtained by training based on stego images generated by ADV-EMB, and ``Natias-SRNet" represents the target steganalytic model SRNet obtained by training based on stego images generated by Natias-ADV. The legends of other figures follow the same pattern. 

The experimental results in Fig.~\ref{retrain} indicate that our method achieves comparable performance with ADV-EMB, whether resisting retrained target steganalytic models or other non-target steganalytic models trained on generated adversarial stego images. When the target model is SRNet and the basic distortion function is SUNIWARD, our method demonstrates a significant improvement, leading to $3.70\%$ average reduction in the detection accuracy of the retrained SRNet. 

Compared with SPS-ENH and USGS, we retrain the target steganalytic model CovNet based on the stego images generated by SPS-ENH, USGS, and our proposed Natias. The relevant experimental results are shown in Table~\ref{retrain-SPS-ENH-cov} and Table~\ref{retrain-USGS-cov}. Regardless of the basic distortion function, our method exhibits a significantly enhanced ability to resist detection by retrained target steganalytic model compared with SPS-ENH. When compared with USGS, our method also achieves comparable performance against retrained steganalytic models.

\begin{table}
\centering
\caption{Detection accuracy ($\%$) of retrained target steganalytic models compared with SPS-ENH. The target steganalytic model is CovNet.}
\label{retrain-SPS-ENH-cov}
\renewcommand{\arraystretch}{1.5}
\begin{tabular}{cccc}
\toprule
Method & Payload & SPS-ENH & Natias-SPS \\ 
\midrule
\multirow{2}{*}{HILL}
& $0.2$ bpp & $70.65$ & $\bm{67.15}$  \\
& $0.4$ bpp & $77.80$ & $\bm{76.25}$  \\
\midrule
\multirow{2}{*}{SUNIWARD} 
& $0.2$ bpp & $70.92$ & $\bm{69.10}$ \\
& $0.4$ bpp & $82.25$ & $\bm{80.85}$  \\         
\bottomrule
\end{tabular}
\end{table}

The experimental results indicate that whether combined with cover enhancement based methods, distortion adjustment based methods or stego post-processing based methods, our method can achieve comparable capability of resisting steganalytic models detection in the retraining scenario.

\subsection{Impact of Different Payloads}
In this section, we discuss the impact of different payloads on transferability performance. According to the experimental results in Table~\ref{trans-ADV} and Table~\ref{trans-USGS-cov}, when our method is combined with ADV-EMB and USGS, the enhancement in transferability performance is restricted in low payload cases. Additionally, for a given target steganalytic model and basic distortion function, as the payload increases, the average gain in transferability performance also increases. We infer that in lower payload cases, the amount of steganographic modification is relatively small, making it difficult to sufficiently corrupt the critical features. Thus, this leads to a restricted enhancement of transferability. Conversely, in higher payload cases, we can thoroughly corrupt them and effectively deceive the target steganalytic model.

However, as illustrated in Table~\ref{trans-SPS-ENH-cov}, although the transferability is also affected by the embedding rate when combined with SPS-ENH, it does not follow the same pattern of transferability change as observed when combined with ADV-EMB and USGS. When combining our method with SPS-ENH, we first add adversarial perturbations to enhance the cover by corrupting critical features, and then add steganographic modifications to the enhanced cover. These steganographic modifications will impact the adversarial perturbations, thereby weakening the corruption of critical features.

\subsection{Impact of Different Target Layers}

In this section, we perform an experimental analysis to illustrate the impact of selecting different target layers on the results. We choose SRNet as the target steganalytic model and select its type $1$ layers, type $2$ layers, type $3$ layers and type $4$ layer as the target layers, respectively. A higher type number corresponds to a deeper layer in SRNet. All experiments are conducted in the $0.4$ bpp case and ADV-EMB is used as the comparative method. The basic distortion function is SUNIWARD. In this section, we utilize the attack success rate (ASR) as a metric, which is defined as the percentage of the stego images that can successfully deceive the target steganalytic model among the total stego images.

\begin{figure}[!t]
\centering
\includegraphics[width=0.8\linewidth]{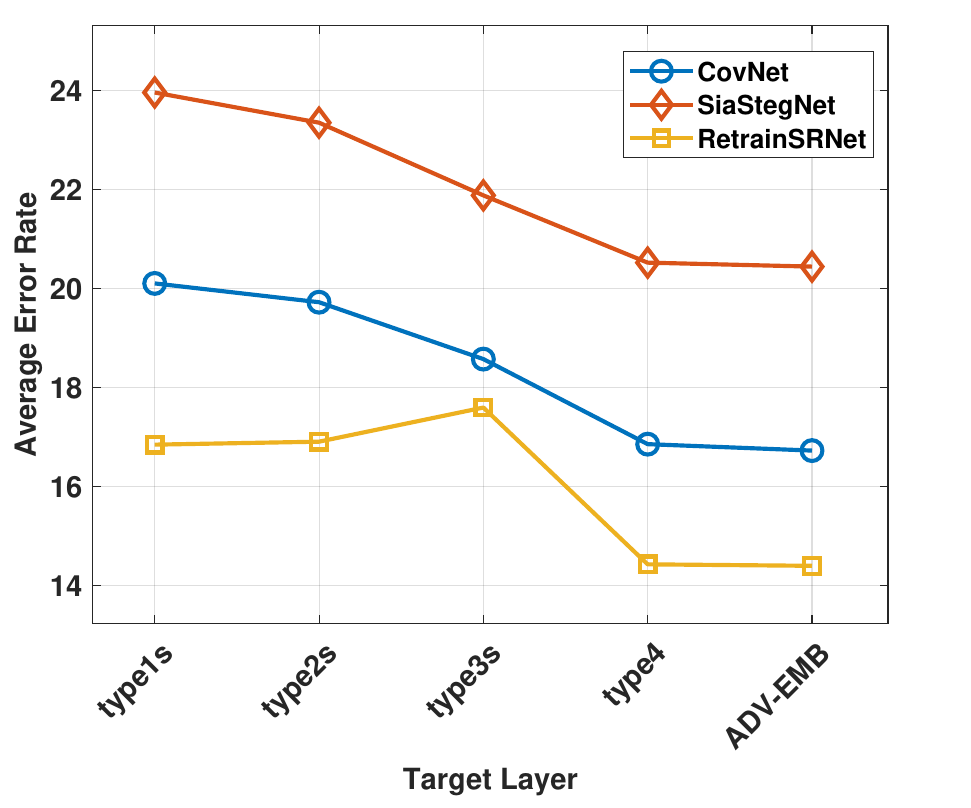}
\caption{Average Error Rate $P_{\text{E}}$ of different steganalytic models when detecting Natias and ADV-EMB under different target layers settings. The target steganalytic model is SRNet. ``RetrainSRNet" represents the new classifier obtained by retraining on the stego images generated by attacking SRNet.}
\label{trans_layer}
\end{figure}

The experimental results of the attack success rate when selecting different target layers are shown in Table~\ref{ASR_layer}. It can be observed that as the selected target layer deepens, the success rate of the attack gradually increases, except for the type $4$ layer. We infer that deeper layers of the network learn more features, while the shallow layers contain low-level features that exert less influence on the output of steganalytic models. However, when selecting the type $4$ layer as the target layer, the size of the corresponding features becomes too narrow, resulting in the loss of valuable information. 
\begin{table}
\centering
\caption{Detection accuracy ($\%$) evaluated on retrained target steganalytic models compared with USGS. The target steganalytic model is CovNet.}
\label{retrain-USGS-cov}
\renewcommand{\arraystretch}{1.5}
\begin{tabular}{cccc}
\toprule
Method & Payload & USGS & Natias-USGS \\ 
\midrule
\multirow{2}{*}{HILL}
& $0.2$ bpp & $\bm{65.30}$ & $65.72$  \\
& $0.4$ bpp & $79.90$ & $\bm{79.15}$  \\
\midrule
\multirow{2}{*}{SUNIWARD} 
& $0.2$ bpp & $68.19$ & $\bm{67.90}$ \\
& $0.4$ bpp & $81.77$ & $\bm{81.15}$  \\         
\bottomrule
\end{tabular}
\end{table}

\begin{table}
\centering
\caption{Attack success rate ($\%$) when choosing different target layers. The target steganalytic model is SRNet. ``Natias-type1s", ``Natias-type2s", ``Natias-type3s" and ``Natias-type4" represent select type $1$ layers, type $2$ layers, type $3$ layers and type $4$ layers as the target layers, respectively.}
\label{ASR_layer}
\renewcommand{\arraystretch}{1.5}
\begin{tabular}{ccc}
\toprule
Adversarial Steganography &  ASR & Feature Size \\ 
\midrule
 ADV-EMB & $93.75\%$  & $[1,1,256,256]$ \\
 Natias-type1s & $84.68\%$  & $[1,16,256,256]$ \\
 Natias-type2s & $86.12\%$  & $[1,16,256,256]$ \\
 Natias-type3s & $94.78\%$  & $[1,256,16,16]$ \\
 Natias-type4 & $93.68\%$  & $[1,512]$ \\
\bottomrule
\end{tabular}
\end{table}

The experimental results of the error rate $P_{\text{E}}$ of different steganalytic models when detecting our methods compared with ADV-EMB under different target layer settings are shown in Fig.~\ref{trans_layer}. Comparing the results in Table~\ref{ASR_layer} and Fig.~\ref{trans_layer}, we can observe that as the layers deepen, the intermediate layer outputs of SRNet become increasingly narrow, and the performance of stego images generated by utilizing different target layers also shows a decreasing trend when resisting detection by non-target steganalytic models including CovNet and SiaStegNet. We infer that as the layers deepen, the interaction between the generated stego images and the target steganalytic model gradually strengthens that leading to overfitting, which results in a reduction in transferability. It means that the adversarial stego images introduce too much adversarial noise to attack the specific features SRNet relying on during steganalysis process.

When selecting the last layer type 4 as the target layer, Natias essentially degrades to ADV-EMB. Because at this point, the final logits are fundamentally attacked rather than the intermediate layer features. When using the retrained SRNet to detect corresponding stego images, the average testing error rate still shows a decreasing trend except when selecting type 3 layers as the target layer. Because the output size of type 3 layers is relatively narrow, as shown in the column ``Feature Size" of Table~\ref{ASR_layer}, and the corresponding features are more concentrative, making it easier to corrupt the critical features and avoid causing dramatic changes to the pixel distribution of the cover image.

Therefore, based on the above experimental results, selecting a narrower layer as the target layer helps us to more effectively corrupt the critical features, deceive various non-target steganalytic models, and enhance the security of our proposed Natias.

\section{Conclusion}\label{conclusion}
In this paper, we propose a novel method Natias to enhance adversarial steganography transferability. Unlike existing adversarial steganographic methods, we first use integrated gradients for neuron attribution to identify critical features. Subsequently, we corrupt these critical features based on the gradient from backpropagation. Finally, we flexibly integrate our approach with various existing adversarial steganographic frameworks to enhance the transferability.

There are still several important issues worth further exploring. For instance, from the perspective of game theory, investigating more theoretically-grounded adversarial steganography methods is a promising research direction. Besides, how to conduct steganalysis specifically targeting the proposed adversarial steganography is also a research question worthy of investigation.


\bibliographystyle{IEEEtran}
\bibliography{nas}

\vspace{11pt}

\begin{IEEEbiography}[{\includegraphics[width=1in,height=1.25in,clip,keepaspectratio]{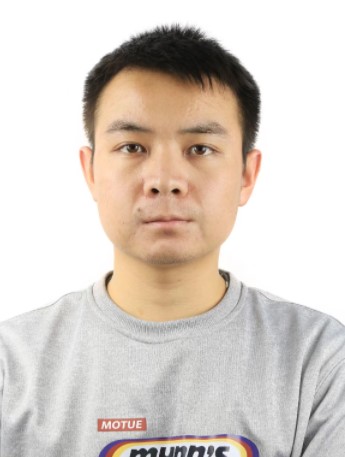}}]{Zexin Fan}
received his B.E. degree in 2021 from the University of Science and Technology of China (USTC). Currently, he is a graduate student at the University of Science and Technology of China. His research interests include adversarial steganography and deep learning.
\end{IEEEbiography}

\begin{IEEEbiography}[{\includegraphics[width=1in,height=1.25in,clip,keepaspectratio]{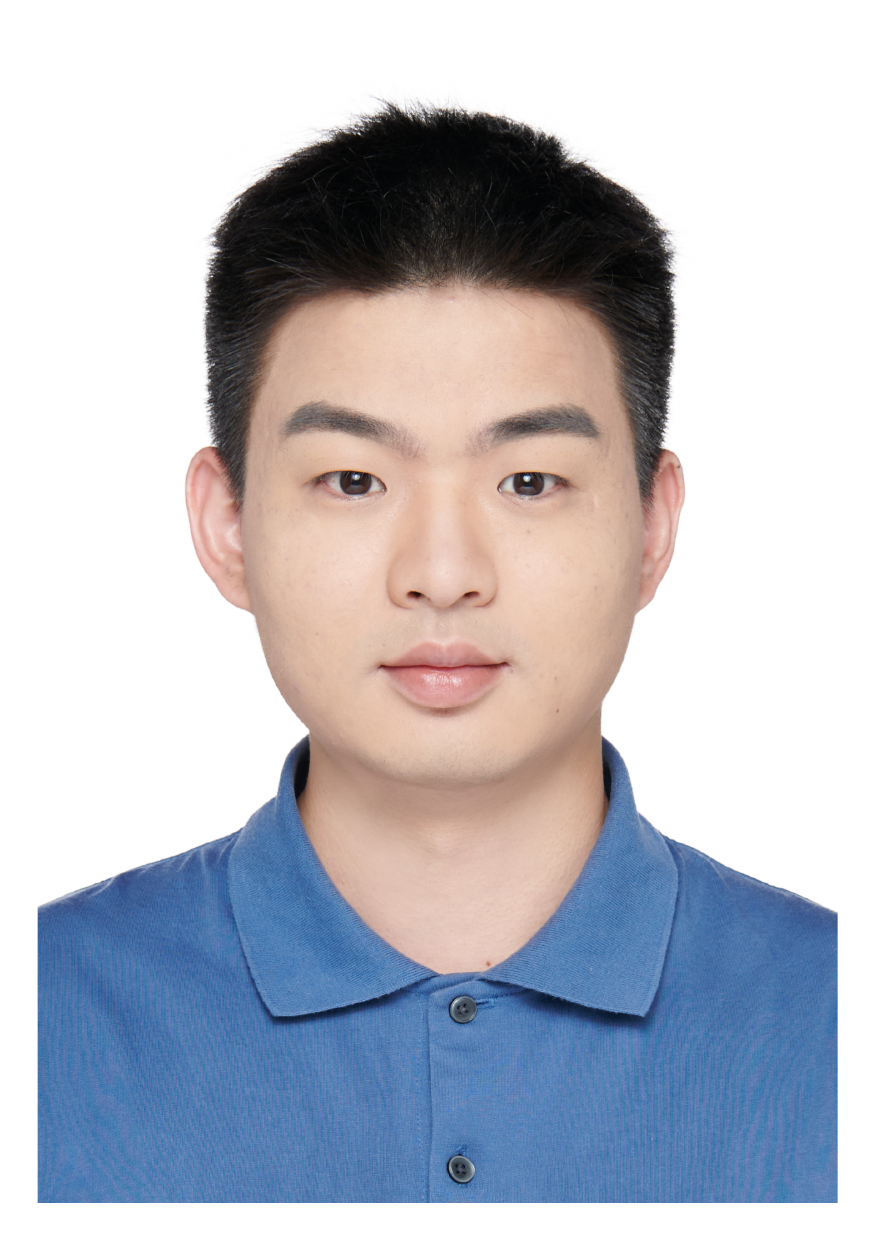}}]{Kejiang Chen}
    received his B.S. degree in 2015 from Shanghai University (SHU) and a Ph.D. degree in 2020 from the University of Science and Technology of China (USTC). Currently, he is an associate research fellow at the University of Science and Technology of China. His research interests include information hiding, image processing and deep learning.
\end{IEEEbiography}

\begin{IEEEbiography}[{\includegraphics[width=1in,height=1.25in,clip,keepaspectratio]{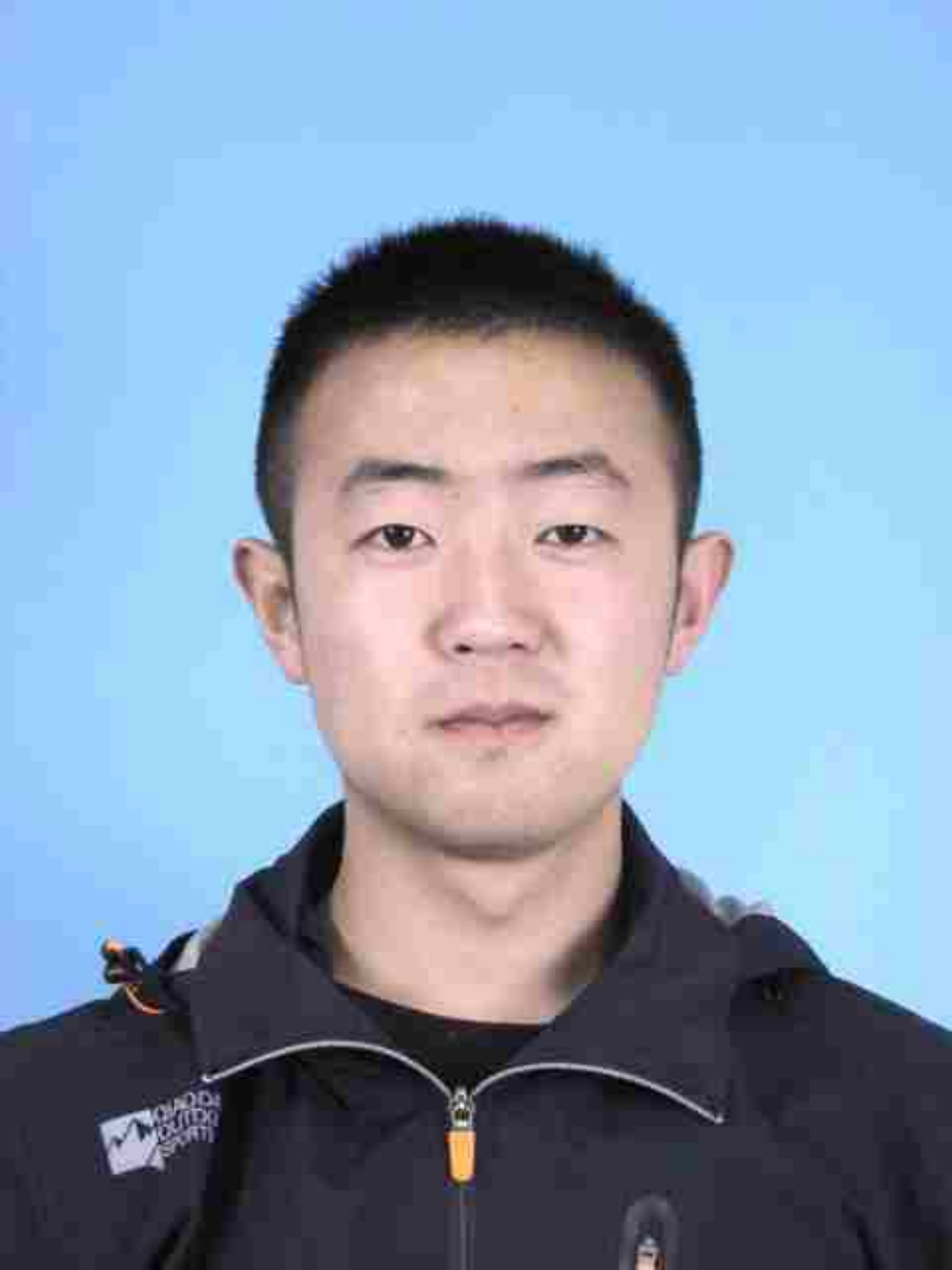}}]{Kai Zeng}
received his B.S. degree in 2019 from the University of Science and Technology of China (USTC). Currently, he is pursuing the
Ph.D. degree with Key Laboratory of Electromagnetic Space Information, School of Information Science and Technology, University of Science and Technology of China, Hefei. His research interests include information hiding and multimedia security.
\end{IEEEbiography}

\begin{IEEEbiography}[{\includegraphics[width=1in,height=1.25in,clip,keepaspectratio]{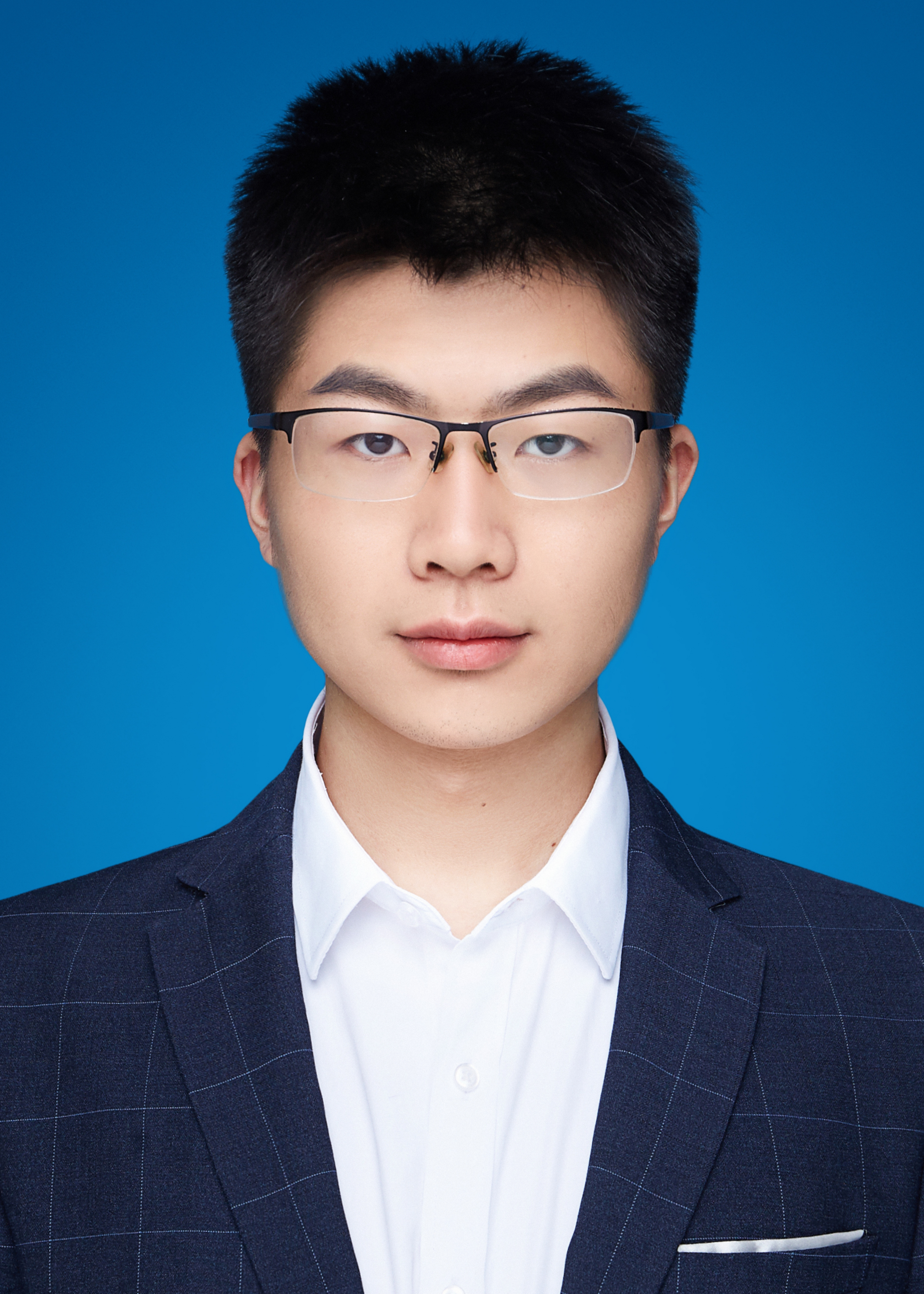}}]{Jiansong Zhang}
    received his B.S. degree in 2019 from the University of Science and Technology of China (USTC). Currently, he is a graduate student at the University of Science and Technology of China. His research interests include steganography, steganalaysis and deep learning.
\end{IEEEbiography}

\begin{IEEEbiography}[{\includegraphics[width=1in,height=1.25in,clip,keepaspectratio]{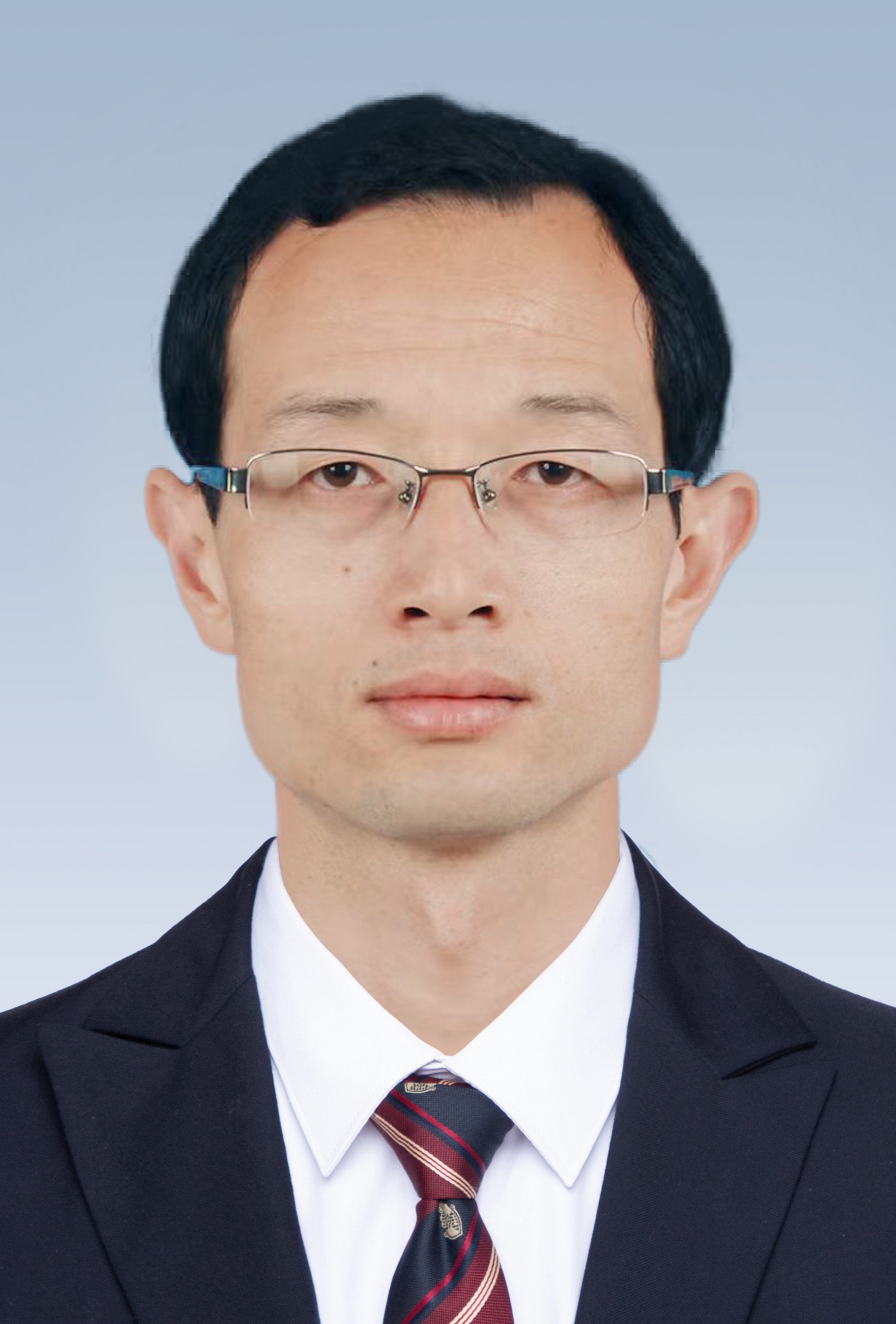}}]{Weiming Zhang}
received his M.S. degree and Ph.D. degree in 2002 and 2005, respectively, from the Zhengzhou Information Science and Technology Institute, P.R. China. Currently, he is a professor with the School of Information Science and Technology, University of Science and Technology of China. His research interests include information hiding and multimedia security.
\end{IEEEbiography}

\begin{IEEEbiography}[{\includegraphics[width=1in,height=1.25in,clip,keepaspectratio]{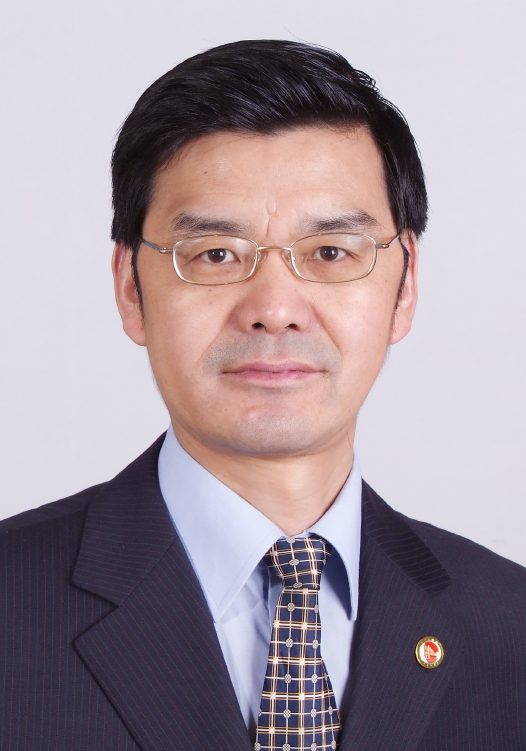}}]{Nenghai Yu}
received his B.S. degree in 1987 from Nanjing University of Posts and Telecommunications, an M.E. degree in 1992 from Tsinghua University and a Ph.D. degree in 2004 from the University of Science and Technology of China, where he is currently a professor. His research interests include multimedia security, multimedia information retrieval, video processing and information hiding.
\end{IEEEbiography}

\end{document}